\documentclass[final,3p,11pt]{elsarticle}

\usepackage{graphicx}
\usepackage{mathptmx} 
\usepackage{latexsym}

\usepackage{amssymb}
\usepackage{lineno}
\usepackage{subfig}
\usepackage{algorithm}
\usepackage[noend]{algpseudocode}
\usepackage{multirow}
\usepackage{booktabs}
\usepackage{multicol}
\usepackage{array}
\usepackage{bm}
\usepackage{amsmath}

\usepackage{hyperref}

\usepackage{natbib}
\biboptions{authoryear}

\usepackage{url}

\usepackage[final,markup=none,commentmarkup=footnote]{changes}
\usepackage{color}
\setcommentmarkup{\footnote{{\bf #2 \arabic{authorcommentcount}:} \ifthenelse{\isColored}{\color{authorcolor}}{}#1}}
\definechangesauthor[color=blue]{YS}
\definechangesauthor[color=maroon]{AE}
\definechangesauthor[color=red]{XL}

\definecolor{marine}{RGB}{0,32,96}
\definecolor{navy}{RGB}{0,0,128}
\definecolor{maroon}{RGB}{128,0,0}
\definecolor{olivegreen}{RGB}{85,107,47}
\definecolor{gray}{RGB}{102,102,102}
\definecolor{green}{RGB}{131,198,210}
\definecolor{blue}{rgb}{0, 0.4470, 0.7410}
\definecolor{skyblue}{rgb}{0.3010, 0.7450, 0.9330}
\definecolor{purple}{rgb}{0.4940, 0.1840, 0.5560}
\definecolor{orange}{rgb}{0.9290, 0.6940, 0.1250}
\definecolor{brown}{RGB}{161,121,124}
\definecolor{red}{rgb}{0.8500, 0.3250, 0.0980}

\usepackage{lipsum}
\makeatletter
\def\ps@pprintTitle{%
	\let\@oddhead\@empty
	\let\@evenhead\@empty
	\def\@oddfoot{}%
	\let\@evenfoot\@oddfoot}
\makeatother

\begin{document}

\begin{frontmatter}
	
	\title{Boosting Ant Colony Optimization via Solution Prediction \\ and Machine Learning}
	
	\author[label1]{Yuan~Sun\corref{cor1}}
	\ead{yuan.sun@monash.edu}
	\cortext[cor1]{Corresponding author}
	\author[label2]{Sheng~Wang}
	\ead{swang@nyu.edu}
	\author[label3]{Yunzhuang~Shen}
	\ead{s3640365@student.rmit.edu.au}
	\author[label3]{Xiaodong Li}
	\ead{xiaodong.li@rmit.edu.au}
	\author[label1]{Andreas~T.~Ernst}
	\ead{andreas.ernst@monash.edu}
	\author[label4]{Michael~Kirley}
	\ead{mkirley@unimelb.edu.au}
	\address[label1]{School of Mathematics, Monash University, Clayton, VIC, 3800,  AU}
	\address[label2]{Center for Urban Science and Progress, New York University, New York, NY, 11201, USA}	
	\address[label3]{School of Computing Technologies, RMIT University, Melbourne, VIC, 3000,  AU}	
	\address[label4]{ School of Computing and Information Systems, The University of Melbourne, Parkville, VIC, 3010, AU}	

\begin{abstract}
This paper introduces an enhanced meta-heuristic (ML-ACO) that combines machine learning (ML) and ant colony optimization (ACO) to solve combinatorial optimization problems. To illustrate the underlying mechanism of our ML-ACO algorithm, we start by describing a test problem, the orienteering problem. In this problem, the objective is to find a route that visits a subset of vertices in a graph within a time budget to maximize the collected score. In the first phase of our ML-ACO algorithm, an ML model is trained using a set of small problem instances where the optimal solution is known. Specifically, classification models are used to classify an edge as being part of the optimal route, or not, using problem-specific features and statistical measures. The trained model is then used to predict the `probability' that an edge in the graph of a test problem instance belongs to the corresponding optimal route. In the second phase, we incorporate the predicted probabilities into the ACO component of our algorithm, i.e., using the probability values as heuristic weights or to warm start the pheromone matrix. Here, the probability values bias sampling towards favoring those predicted `high-quality' edges when constructing feasible routes. We have tested multiple classification models including graph neural networks, logistic regression and support vector machines, and the experimental results show that our solution prediction approach consistently boosts the performance of ACO. Further, we empirically show that our ML model trained on small synthetic instances generalizes well to large synthetic and real-world instances. Our approach integrating ML with a meta-heuristic is generic and can be applied to a wide range of optimization problems.
\end{abstract}


\end{frontmatter}

\section{Introduction}\label{Section: Introduction}
Ant colony optimization (ACO) is a class of widely-used meta-heuristics, inspired by the foraging behavior of biological ants, for solving combinatorial optimization problems \citep{dorigo1996ant,dorigo1997ant}. Since its introduction in early 1990s, ACO has been extensively investigated to understand both its theoretical foundations and practical performance \citep{dorigo2005ant,blum2005ant}. A lot of effort has been made to improve the performance of ACO, making it one of the most competitive algorithms for solving a wide range of  optimization problems. Whilst ACO cannot provide any optimality guarantee due to its heuristic nature, it is usually able to find a high-quality solution for a given problem within a limited computational budget. 

The ACO algorithm builds a probabilistic model to sample solutions for an optimization problem. In this sense, ACO is closely related to Estimation of Distribution Algorithms and Cross Entropy methods \citep{zlochin2004model}. The probabilistic model of ACO is parametrized by a so-called \emph{pheromone matrix} and a \emph{heuristic weight matrix}, which basically measure the `payoff' of setting a decision variable to a particular value. The aim of ACO is to evolve the pheromone matrix so that an optimal (or a  near-optimal) solution can be generated via the probabilistic model in sampling. Previously, the pheromone matrix is usually initialized uniformly and the heuristic weights are set based on prior domain knowledge. In this paper, we develop machine learning (ML) techniques to warm start the pheromone matrix or identify good heuristic weights for ACO to use.

Leveraging ML to help combinatorial optimization has attracted much attention recently \citep{bengio2018machine,karimi2021machine}. For instance, novel ML techniques have been developed to prune the search space of large-scale optimization problems to a smaller size that is manageable by existing solution algorithms \citep{sun2019using,lauri2019fine,sun2020generalization}, to order decision variables for branch and bound or tree search algorithm \citep{li2018combinatorial,shen2021learning}, \added[]{and to approximate the objective value of solutions \citep{fischetti2019machine,santini2021probabilistic}.} There also exist ML-based methods that try to directly predict a high-quality solution for an optimization problem \citep{abbasi2020predicting,ding2019accelerating}. The key idea of these methods is typically \emph{solution prediction} via ML; that is aiming to predict the optimal solution for a given problem as close as possible. 

Building upon these previous studies, we propose an enhanced meta-heuristic named ML-ACO, that combines ML (more specifically solution prediction) and ACO to solve combinatorial optimization problems. To illustrate the underlying mechanism of our proposed algorithm, we first describe the orienteering problem, which is also used to demonstrate the efficacy of ML-ACO. The aim of orienteering problem is to search for a route in a graph that visits a subset of vertices within a given time budget to maximize the total score collected from the visited vertices (see Section \ref{Subsection: Orienteering Problem} for a formal definition). The orienteering problem has many real-world applications \citep{vansteenwegen2011orienteering,gunawan2016orienteering}.

\begin{figure*}[!t]
	\centering
	\includegraphics[scale=0.9]{ 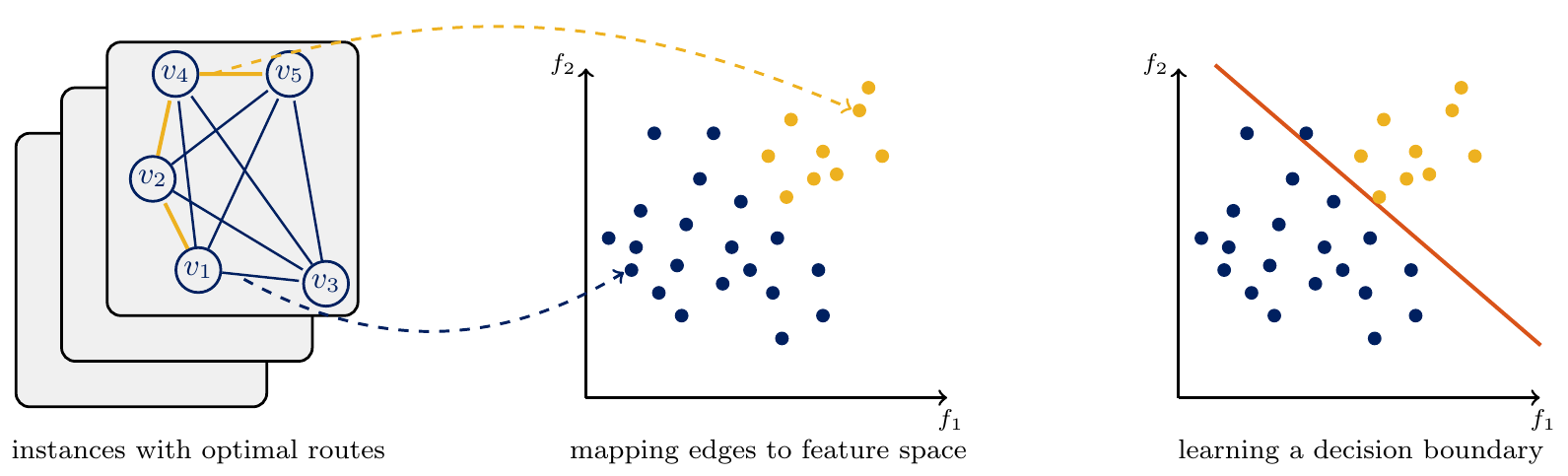}
	\caption{An illustration of the training procedure of our ML model. First, a set of orienteering problem instances are solved, with the optimal route highlighted  in yellow in the corresponding graph of    orienteering problem instance (left figure). We then extract features (e.g., edge weight) to describe each edge of the graphs, and map each edge to the feature space as a training point (middle figure). Finally, we apply a classification algorithm to learn a decision boundary in the feature space to well separate edges (training points) that are part of the optimal routes from those which are not (right figure).}
	\label{Fig: An illustration of the training procedure of our ML model}
\end{figure*}

\begin{figure*}[!t]
	\centering
	\includegraphics[scale=0.9]{ 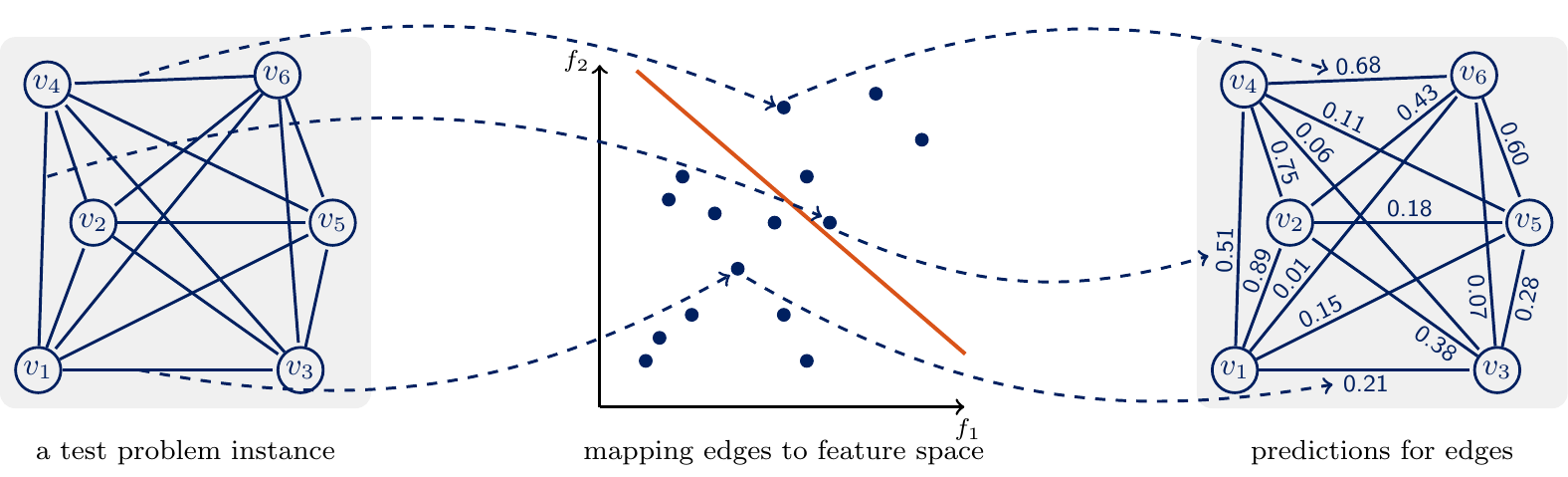}
	\caption{The testing procedure of our ML  model. Given an unsolved test orienteering problem instance (left figure), we first map each edge of the corresponding graph to a point in the feature space (middle figure). Based on the location of the points with respect to the decision boundary learned in training, we can compute for each edge a probability that it belongs to an optimal route (right figure). The predictions are then used to bias the sampling of ACO towards using the edges with a larger predicted probability value more often when constructing feasible routes.}
	\label{Fig: An illustration of the testing procedure of our ML model}
\end{figure*}

In the first phase of our ML-ACO algorithm,  an ML model is trained on a set of optimally-solved small orienteering problem instances with known optimal route, as shown in Figure~\ref{Fig: An illustration of the training procedure of our ML model}. We extract problem-specific features as well as statistical measures (see Section \ref{Subsection: Constructing Training Set}) to describe each edge in the graphs of solved orienteering problem instances, and map each edge to a training point in the feature space. Classification algorithms can then be used to learn a decision boundary in the feature space to differentiate the edges that are in the optimal routes from those which are not. We have tested multiple existing classification algorithms for this task including graph neural networks \citep{kipf2016semi,wu2020comprehensive}, logistic regression \citep{bishop2006pattern} and support vector machines \citep{boser1992training,cortes1995support}. For an unsolved test orienteering problem instance, the trained ML model can then be used to predict the `probability' that an edge in the corresponding graph belongs to the optimal route, as shown in Figure~\ref{Fig: An illustration of the testing procedure of our ML model}. 

In the second phase of our ML-ACO algorithm, we incorporate the probability values predicted by our ML model into the ACO algorithm, i.e., using the  probability values to set the \emph{heuristic weight matrix} or to initialize the \emph{pheromone matrix} of ACO. The aim is to bias the sampling of ACO towards favoring the edges that are predicted more likely to be part of an optimal route, and hopefully to improve the efficiency of ACO in finding high-quality routes. In this sense, the idea of our ML-ACO algorithm is also related to the \emph{seeding} strategies that are used to improve evolutionary algorithms \citep{liaw2000hybrid,hopper2001empirical,friedrich2015seeding,chen2018effects}.

We use simulation experiments to show the efficacy of our ML-ACO  algorithm on the orienteering problem. The results show that our ML-ACO algorithm significantly improves over the classic ACO in finding high-quality solutions. We also test the use of different classification algorithms, and observe that our solution prediction approach consistently boosts the performance of ACO. Finally, we show that our ML model trained on small synthetic problem instances generalizes well to large synthetic instances as well as real-world instances. 


In summary, we have made the following contributions:
\begin{itemize}
	\item This paper is the first attempt, to our knowledge, at boosting the performance of the ACO algorithm via solution prediction and ML. 
	\item We empirically show that our proposed ML-ACO algorithm significantly improves over the classic ACO, no matter which classification algorithm is used in training. 
	\item We also demonstrate the generalization capability of  ML-ACO on large synthetic and real-world problem instances. 
\end{itemize}

The remainder of this paper is organized as follows. In Section \ref{Section: Ant Colony Optimization for Solving Orienteering Problem}, we introduce the orienteering problem and the ACO algorithm. In Section \ref{Section: Boosting ACO via Solution Prediction}, we describe the proposed ML-ACO algorithm. Section \ref{Section: Experiments} presents our experimental results, and the last section concludes the paper and shows potential avenues for future research.

\section{Background and Related Work}\label{Section: Ant Colony Optimization for Solving Orienteering Problem}
We first describe the orienteering problem and then introduce the ACO algorithm in the context of orienteering problem. 

\subsection{Orienteering problem}\label{Subsection: Orienteering Problem}


The orienteering problem finds its application in many real-world problems, such as tourist trip planning, home fuel delivery and building telecommunication networks \citep{vansteenwegen2011orienteering,gunawan2016orienteering}. \added[]{Consider a complete undirected graph $G(V,E,S,C)$, where $V=\{v_1,\ldots, v_n\}$ denotes the vertex set, $E=\{e_{i,j} \, | \,  1\le i \ne j\le n \}$ denotes the edge set, $S=\{s_1,\ldots,s_n\}$ denotes the score of each vertex, and $C = \{c_{i,j} \, | \,  1\le i \ne j\le n \}$ denotes the time cost of traveling through each edge.} Assume $v_1$ is the starting vertex and $v_n$ is the ending vertex. The objective of the orienteering problem is to search for a path \added[]{from $v_1$ to $v_n$} that visits a subset of vertices within a given time budget $T_\mathrm{max}$, such that the total score collected is maximized. Thus, the orienteering problem can be viewed as a combination of traveling salesman problem and knapsack problem. We use $u_i$ to denote the visiting order of vertex $v_i$,  and use a binary variable $x_{i,j}$ to denote whether vertex $v_j$ is visited directly after vertex $v_i$. The integer program of the orienteering problem can be written as:
\begin{align}
\max_{\bm{x}, \bm{u}} & \sum_{i=1}^{n-1} \sum_{j=2}^{n} s_{j} x_{i,j}, \\
s.t. \;\, & \sum_{j=1}^{n} x_{1,j} = \sum_{i=1}^{n} x_{i,n} = 1, \label{Eq. con1}\\
& \sum_{i=1}^{n-1} x_{i,k}  = \sum_{j=2}^{n} x_{k,j} \le 1, \; && 2 \le k \le n-1; \label{Eq. con2}\\
& u_i - u_j + 1 \le (n - 1)(1-x_{i,j}), && 2 \le i, j \le n; \label{Eq. con3} \\
& \sum_{i=1}^{n-1}  \sum_{j=2}^{n}  c_{i,j} x_{i,j} \le T_\mathrm{max}, \label{Eq. con4} \\
&  u_i \ge 0, \;&& 2 \le i \le n; \\
& x_{i,j} \in \{0, 1\}, \; &&1 \le i, j \le n.
\end{align}
The \added[]{constraints} (\ref{Eq. con1}) ensure that the path starts from vertex $1$ and ends in vertex $n$. The constraints (\ref{Eq. con2}) guarantee that each vertex in between can only be visited at most once. The constraints (\ref{Eq. con3}) are the Miller-Tucker-Zemlin subtour elimination constraints, and the constraint (\ref{Eq. con4}) satisfies the given time budget. Note that this formulation is not computationally efficient, and there are some relatively trivial ways to make it slightly stronger \added[]{\citep{fischetti1998solving}}. However, this formulation is sufficient for logical correctness. 

The orienteering problem is NP-hard \citep{golden1987orienteering}. Many solution methods have been proposed to solve the orienteering problem and its variants, including exact solvers \citep{fischetti1998solving, el2016solving, archetti2016branch,angelelli2017probabilistic} and heuristics or meta-heuristics \citep{kobeaga2018efficient,santini2019adaptive,hammami2020hybrid,assunccao2021coupling}. Solving the orienteering problem to optimality using exact solvers may take a long time, especially for large instances. However, in some real-world applications such as tourist trip planning, we need to provide a high-quality solution to users within a short time. In this case, meta-heuristics are useful to search for a high-quality solution when the computational budget is very limited.  In the next subsection, we describe the meta-heuristic, ACO, to solve the orienteering problem.

\subsection{Ant colony optimization}\label{Subsection: Ant Colony Optimization}

The ACO algorithm is inspired by the behavior of biological ants seeking the shortest path between food and their colony \citep{dorigo1996ant,dorigo1997ant}. Unlike many other nature inspired algorithms, ACO has a solid mathematical foundation, based on probability theory. The underlying mechanism of ACO is to build a parametrized probabilistic model to incrementally construct feasible solutions. The parameters of this probabilistic model are evolved over time, based on the sample solutions generated in each iteration of the algorithm. By doing this, better solution components are reinforced, leading to an optimal (or near-optimal)  solution in the end. The ACO algorithm has been demonstrated to be effective in solving various combinatorial optimization problems \citep{dorigo2005ant, blum2005ant, mavrovouniotis2016ant, xiang2021pairwise,jia2021bilevel,palma2021optimised}. 

As our main focus is to investigate whether ML can be used to improve the performance of ACO, we simply test on two representative ACO models -- Ant System (AS) \citep{dorigo1996ant} and Max-Min Ant System (MMAS) \citep{stutzle2000max}. The AS is one of the original ACO algorithms, and MMAS is a \added[]{well-performing} variant \citep{blum2005ant}. Note that ACO has been applied to solve the orienteering problem variants, e.g.,  team orienteering problem \citep{ke2008ants}, team orienteering problem with time windows \citep{montemanni2011enhanced,gambardella2012coupling} and time-dependent orienteering problem with time windows \citep{verbeeck2014fast,verbeeck2017time}. These works are typically based on one of the early ACO models, possibly integrated with local search methods. To avoid complication, we simply select the AS and MMAS models, which are sufficient for our study. 

The \textbf{AS} algorithm uses a population of $m$ ants to incrementally construct feasible solutions based on a parametrized probabilistic model. For the orienteering problem, a feasible solution is a path, consisting of a set of connected edges. In one iteration of the algorithm, each of the $m$ ants constructs its own path from scratch. Starting from $v_1$, an ant incrementally selects the next vertex to visit until all the time budget is used up. Note that as $v_n$ is the ending vertex, each ant should reserve enough time  to visit $v_n$. 

Suppose an ant is at vertex $v_i$, and $V_i$ denotes the set of vertices that this ant can visit in the next step without violating the time budget constraint. The probability of this ant visiting vertex $v_j \in V_i$ in the next step is defined by 
\begin{equation}\label{Eq: probabilistic model}
p_{i,j} = \frac{\tau_{i,j}^\alpha\eta_{i,j}^\beta}{\sum\limits_{k\in V_i}\tau_{i,k}^\alpha\eta_{i,k}^\beta}, 
\end{equation}
where $\tau_{i,j}$ is the amount of \emph{pheromone} deposited by the ants for transition from vertex $v_i$ to $v_j$; $\eta_{i,j}$ is the \emph{desirability of transition} from vertex $v_i$ to $v_j$; $\alpha \ge 0$ and $\beta \ge 0$ are control parameters. This means the probability of visiting vertex $v_j \in V_i$  from $v_i$ is proportional to the product of $\tau_{i,j}^\alpha\eta_{i,j}^\beta$. 

The desirability of transition from vertex $v_i$ to $v_j$, i.e., $\eta_{i,j}$, is usually set based on prior knowledge. In the orienteering problem, we can set $\eta_{i,j}$ to the ratio of the score collected at vertex $v_j$ to the time required for travelling from vertex $v_i$ to $v_j$: $\eta_{i,j} = s_{j}/c_{i,j}$. This computes the score that can be collected per unit time if travelling through edge $e_{i,j}$, and measures the `payoff' of including edge $e_{i,j}$ in the path in terms of the objective value. By using these $\bm{\eta}$ values, high-quality edges (i.e., those allowing for a large collected score per unit time) are more likely to be sampled.  

The pheromone values $\bm{\tau}$ are typically initialized uniformly, and are gradually evolved in each iteration of the algorithm, such that the components (edges) of high-quality sample solutions gradually acquire a large pheromone value. This also biases the sampling to using high-quality edges more often. In each iteration, after all the $m$ ants have completed their solution construction process, the pheromone values $\tau_{i,j}$, where $i, j = 1, \ldots, n$ and $i\ne j$, are  updated based on the sample solutions generated:
\begin{equation}
\tau_{i,j} = (1-\rho) \tau_{i,j} + \sum_{k=1}^{m}\Delta\tau_{i,j}^k, 
\end{equation}
where $\rho>0$ is the \emph{pheromone evaporation coefficient}, and $\Delta\tau_{i,j}^k$ is the amount of pheromone deposited by the $k^\mathrm{th}$ ant on edge $e_{i,j}$. Let $y_k$ denote the objective value collected by the $k^\mathrm{th}$ ant, and $C>0$ be a  constant. We can define  $\Delta\tau_{i,j}^k = y_k/C$, if edge $e_{i,j}$ is used by the $k^\mathrm{th}$ ant; otherwise $\Delta\tau_{i,j}^k = 0$. The amount of pheromone deposited by an ant when it travels along a path is proportional to the objective value of the path. As we are solving a maximization problem, edges that appear in high-quality paths are reinforced (i.e., acquiring a large pheromone value), so that these edges are more likely to be used when constructing paths in the later iterations. This sampling and evolving process is repeated for a predetermined number of iterations, and the best solution generated is returned in the end. 

The \textbf{MMAS} algorithm is a variant of AS, which uses the same probabilistic model (Eq. \ref{Eq: probabilistic model}) to construct feasible solutions. The key difference between MMAS and AS is how the pheromone matrix ($\bm{\tau}$) is updated. The MMAS algorithm only uses a single solution to update the pheromone matrix in each iteration, in contrast to AS which uses a population of $m$ solutions. This single solution can either be the best solution generated in the current iteration (\emph{iteration-best}) or the best one found so far (\emph{global-best}). The use of a single best solution makes the search more greedy towards high-quality solutions, in the sense that only the edges in the best solution get reinforced. Let $\bm{x}^\mathrm{best}$ denote the best solution  and $y^\mathrm{best}$ be the objective value of $\bm{x}^\mathrm{best}$. The pheromone values $\tau_{i,j}$ for each pair of $i,j=1,\ldots,n$ and $i \ne j$ are updated as 
\begin{equation}\label{Eq. Pheromone update MMAS}
\tau_{i,j} = (1-\rho) \tau_{i,j} + \Delta\tau_{i,j}^\mathrm{best}, 
\end{equation}
where $\Delta\tau_{i,j}^\mathrm{best} = 1/y^\mathrm{best}$ if edge $e_{i,j}$ is in the best solution $\bm{x}^\mathrm{best}$; otherwise $\Delta\tau_{i,j}^\mathrm{best} = 0$. 

The second key difference between MMAS and AS is that the pheromone values in MMAS are restricted to a range of $[\tau_\mathrm{min}, \tau_\mathrm{max}]$. After the pheromone values have been updated in each iteration using Eq. (\ref{Eq. Pheromone update MMAS}), if a pheromone value $\tau_{i,j} > \tau_\mathrm{max}$, we reset it to the upper bound $\tau_\mathrm{max}$. This avoids the situation where  an edge accumulates a very large pheromone value, such that it is (almost) always selected in sampling  based on the probabilistic model. The upper bound on pheromone values is derived as 
\begin{equation}\label{Eq: tmax}
	\tau_\mathrm{max} = \frac{1}{\rho \cdot y^\mathrm{opt}},
\end{equation}
where $y^\mathrm{opt}$ is the optimal solution of the problem instance. In practice, we often substitute $y^\mathrm{opt}$ with the best solution found so far to compute the upper bound, since we do not have  $y^\mathrm{opt}$ before solving the problem. Similarly if a pheromone value $\tau_{i,j} < \tau_\mathrm{min}$, we reset it to the lower bound $\tau_\mathrm{min}$. This ensures the probability of selecting any edge in sampling does not reduce to zero. In this sense, the probability of generating the optimal solution via sampling approaches one if given an infinite amount of time. We simply set the lower bound to 
\begin{equation}\label{Eq: tmin}
	\tau_\mathrm{min} = \frac{\tau_\mathrm{max}}{2n},
\end{equation}
 where $n$ is the problem dimensionality. \added[]{This setting is consistent with the original paper \citep{stutzle2000max} for solving the traveling salesman problem.}
 
 The MMAS algorithm also uses an additional mechanism called \emph{pheromone trail smoothing} to deal with premature convergence. When the algorithm has converged, the pheromone values are increased proportionally to their difference to $\tau_\mathrm{max}$:
 \begin{equation}
 \tau_{i,j} = \tau_{i,j} + \delta \cdot (\tau_\mathrm{max} - \tau_{i,j}), 
 \end{equation}
 where $0 \le \delta \le 1$ is a control parameter. In the extreme case  $\delta = 1$, it is equivalent to \added[]{restarting} the algorithm, in the sense that the pheromone values are reinitialized uniformly to $\tau_\mathrm{max}$. We activate the \emph{pheromone trail smoothing} mechanism if there is no improvement in the objective value for a predetermined number of consecutive iterations $T_\mathrm{pts}$.

\section{Boosting Ant Colony Optimization via Solution Prediction}\label{Section: Boosting ACO via Solution Prediction}
This section presents the proposed ML-ACO algorithm, that integrates ML with ACO to solve the orienteering problem. The main idea of our ML-ACO algorithm is first to develop an ML model, aiming to predict the probability that an edge in the graph of an orienteering problem instance belongs to the optimal route. The training and testing procedures of our ML model are illustrated in Figure~\ref{Fig: An illustration of the training procedure of our ML model} and \ref{Fig: An illustration of the testing procedure of our ML model} respectively. The predicted probability values are then leveraged to improve the performance of ACO in finding high-quality solutions.  

In the first phase of our ML-ACO algorithm,  we construct a training set from small orienteering problem instances (graphs), that are solved to optimality by a generic exact solver -- CPLEX. We treat each edge in a solved graph as a training point, and extract several graph features as well as statistical measures to characterize each edge (Section \ref{Subsection: Constructing Training Set}). The edges that are part of the optimal route obtained by CPLEX are called \emph{positive} training points and labelled as $1$; otherwise they are \emph{negative} training points labelled as $-1$. Note that the decision variables of the edges that do not belong to the optimal route have a value of zero in the optimal solution produced by CPLEX. We call these edges \emph{negative} training points to be consistent with the ML literature. This then becomes a binary classification problem, where the goal is to learn a decision rule based on the extracted features to well separate the positive and negative training points. We will test multiple classification algorithms for this task. Given an \emph{unsolved} orienteering problem instance, the trained ML model can then be applied to predict for each edge a probability that it belongs to the optimal route (Section \ref{Subsection: Training and Solution Prediction}). 

In the second phase of ML-ACO, the probability values predicted by our ML model are then incorporated into the probabilistic model of ACO to improve its performance. The idea is to use the edges that are predicted more likely to be in an optimal route more often in the sampling process of ACO. By doing this, high-quality routes can hopefully be generated more quickly. We will use the predicted probability values to either seed the \emph{pheromone matrix} or set the \emph{heuristic weight matrix} of ACO (Section \ref{Subsection: Incorporating Solution Prediction into ACO}).

The general procedure of our ML-ACO algorithm can be summarized as follows:
\begin{enumerate}
	\item Solve small orienteering problem instances to optimality using CPLEX;
	\item Construct a training set from the optimally-solved problem instances;
	\item Train an ML model offline to separate positive and negative training points (edges) in our training set; 
	\item Predict which edges are more likely to be in an optimal route for a test (unsolved) problem instance;
	\item Incorporate solution prediction into ACO to boost its performance. 	
\end{enumerate}

Note that our ML model is based on the edge representation of solutions for the orienteering problem, i.e., each edge in the graph belongs to a route (solution) or not. A more efficient way typically used by ACO to represent a route is using a sequence of vertices. A potential avenue for future research would be to develop an ML model based on the vertex representation to predict the order in which vertices are visited in the optimal route. This will become a \emph{multiclass} classification problem in contrast to the \emph{binary} classification problem developed in this paper.

\subsection{Constructing training set}\label{Subsection: Constructing Training Set}

We construct a training set from optimally-solved orienteering problem instances on complete graphs, where each edge is a training point.  We assign a class label $1$ to edges that belong to the optimal route and $-1$ to those who do not. Three graph features and two statistical measures are designed to characterize each edge, which are detailed below. 

Recall that the objective of orienteering problem $G(V,E,S,C)$ is to search for a path that visits a subset of vertices within a given time budget $T_\mathrm{max}$, to maximize the total score collected. Three factors are relevant to the objective, i.e., vertex scores $S$, edge costs $C$ and time budget $T_\mathrm{max}$. The first graph feature we design to describe edge $e_{i,j}$, is the ratio between edge cost $c_{i,j}$ and the time budget $T_\mathrm{max}$
\begin{equation}
f_1(e_{i,j}) = \frac{c_{i,j}}{T_\mathrm{max}},
\end{equation} 
where $i, j = 1,\ldots, n$ and $i \ne j$. Intuitively, if $c_{i,j} > T_\mathrm{max}$, the edge $e_{i,j}$ certainly cannot appear in any of the feasible solutions. A stronger preprocessing criterion would be to eliminate edge $e_{i,j}$ if \added[]{$c_{1,i} + c_{i,j} + c_{j,n} > T_\mathrm{max}$}, where $v_1$ is the starting vertex and $v_n$ is the ending vertex. However, this type of exact pruning mechanism is not expected to eliminate many edges from a problem instance.  

Another informative feature for describing edge $e_{i,j}$ is the ratio between vertex score $s_j$ and edge cost $c_{i,j}$, which computes the score we can collect immediately from vertex $v_j$ per unit time if taking the edge $e_{i,j}$. We normalize this ratio of edge $e_{i,j}$ by the maximum ratio of the edges that originates from vertex $v_i$,
\begin{equation}
f_2(e_{i,j}) = \frac{s_{j}/c_{i,j}}{\max\limits_{k=1,\ldots, n} s_{k} / c_{i,k}}.
\end{equation}
This normalization is useful because it computes the \emph{relative} payoff of selecting edge $e_{i,j}$, comparing to the alternative ways of leaving vertex $v_i$. Similarly, we also normalize the ratio of edge $e_{i,j}$ by the maximum ratio of the edges that ends in vertex $v_j$, 
\begin{equation}
f_3(e_{i,j}) = \frac{s_{j}/c_{i,j}}{\max\limits_{k=1\ldots,n} s_{j} / c_{k,j}}. 
\end{equation} 
This computes the \emph{relative} payoff of visiting vertex $v_j$ via edge $e_{i,j}$,  comparing against other ways of visiting vertex $v_j$. These graph features are computationally very cheap, but they only capture local characteristics of an edge. Hence, we also adopt two statistical measures, originally proposed in \citep{sun2019using}, to capture global features of an edge. 

The two statistical measures rely on random samples of feasible solutions (routes). We use the method presented in Appendix~\ref{Subsection: Random Sampling Method for Orienteering Problem} to generate $m$ random feasible solutions, denoted as $\{\bm{x}^1,\bm{x}^2,\ldots,\bm{x}^m\}$, and their objective values denoted as $\{y^1,y^2,\ldots,y^m\}$. Each solution $\bm{x}$ is a binary string, where $x_{i,j} = 1$ if the edge $e_{i,j}$ is in the route; otherwise $x_{i,j} = 0$. The time complexity of sampling $m$ feasible solutions for an $n$-dimensional problem instance is $\mathcal{O}(mn)$, which is proved in Appendix~\ref{Subsection: Random Sampling Method for Orienteering Problem}. The sample size $m$ should be larger than $n$, otherwise there will be some edges that are never sampled. \added[]{We will set $m = 100n$ in our experiments, unless explicitly indicated otherwise.}

The first statistical measure for characterizing edge $e_{i,j}$ is computed based on the ranking of sample solutions
\begin{equation}\label{Eq: ranking-based measure}
f_r(e_{i,j}) = \sum_{k=1}^{m}\frac{x_{i,j}^k}{r^k},
\end{equation}
where $r^k$ denotes the ranking of the $k^\mathrm{th}$ sample in terms of its objective value in descending order. This ranking-based  measure assigns a large score to edges that \emph{frequently} appear in \emph{high-quality} sample solutions, in the hope that these edges may also appear in an optimal solution. We normalize the ranking-based score of each edge by the maximum score in a problem instance to alleviate the effects of different sample size $m$
\begin{equation}
f_4(e_{i,j}) =\frac{f_r(e_{i,j})}{\max\limits_{p,q = 1,\ldots,n} f_r(e_{p,q})}. 
\end{equation}

The other statistical measure employed is a correlation-based measure, that computes the Pearson correlation coefficient between each variable $x_{i,j}$ and objective values $y$ across the sample solutions:
\begin{equation}\label{Eq: correlation-based measure}
f_c(e_{i,j}) = \frac{\sum_{k=1}^{m}(x_{i,j}^k - \bar{x}_{i,j})(y^k - \bar{y}) }{\sqrt{\sum_{k=1}^{m}(x_{i,j}^k-\bar{x}_{i,j})^2}\sqrt{\sum_{k=1}^{m}(y^k-\bar{y})^2}},
\end{equation}
where $\bar{x}_{i,j} = \sum_{k= 1}^{m}x_{i,j}^k/m$, and $\bar{y} = \sum_{k=1}^{m}y^k/m$. As the orienteering problem is a maximization problem, edges that are highly positively correlated with the objective values are likely to be in an optimal route. Similarly, we normalize the correlation-based score of each edge by the maximum correlation value in a problem instance:
\begin{equation}
f_5(e_{i,j}) = \frac{f_c(e_{i,j})}{\max\limits_{p,q = 1, \ldots, n} f_c(e_{p,q})}. 
\end{equation}

The time complexity of directly computing these two statistical measures based on the binary string representation $\bf{x}$ is $\mathcal{O}(mn^2)$, as we need to visit every bit in each of the $m$ binary strings. To improve the time efficiency, we adopt the method proposed in \citep{sun2019using}, which represents the sample solutions using sets instead of strings. We then are able to compute the statistical measures in $\mathcal{O}(mn + n^2)$ time. The details of how to efficiently compute these measures are presented in Appendix~\ref{Subsection: Computing Statistical Measures Efficiently}. 

In summary, we have extracted five features ($f_1$, $f_2$, $f_3$, $f_4$, $f_5$) to characterize each edge (training point). For a problem instance with $n$ vertices, we can extract $n(n-1)$ training points, as there are $n(n-1)$ directed edges in the corresponding complete graph. We use multiple solved problem instances to construct our training set $\mathbb{S}=\{(\bm{f}^i,l^i)\,|\, i = 1,\ldots,n_t\}$, where $\bm{f}^i$ is the $5$-dimensional feature vector;  $l^i\in\{-1 , 1\}$ is the class label of the $i^\mathrm{th}$ training point; and $n_t$ is the  number of training points.

\subsection{Training and solution prediction}\label{Subsection: Training and Solution Prediction}

After we have obtained a training set $\mathbb{S}$, our goal is then to learn a decision boundary to separate positive (label $1$) and negative (label $-1$) training points in $\mathbb{S}$ as well as possible. This is a typical binary classification problem, that can be solved by any off-the-shelf classification algorithm. To see the effects of using different classification algorithms, we compare three alternatives for this task, namely, support vector machine (SVM) \citep{boser1992training,cortes1995support}, logistic regression (LR)  \citep{bishop2006pattern}, and graph convolutional network (GCN) \citep{kipf2016semi,wu2020comprehensive}. SVM and LR are well-known traditional algorithms with a solid mathematical foundation, and GCN is a popular deep neural network based on graph structure of a problem. This comparison is interesting, because it sheds light on whether a `deep' model outperforms a `shallow' model in the context of solution prediction for combinatorial optimization. A brief description of the three learning algorithms can be found in Appendix~\ref{Subsection: ml algorithms}. 

In our training set, the number of positive training points is much smaller than that of negative training points. Considering an orienteering problem instance with $n$ vertices, the number of edges \added[]{appearing} in an optimal route is less than $n$, and the total number of edges in the directed complete graph is $n(n-1)$. Hence, the ratio between positive and negative edges is less than $1/(n-2)$. In this sense, our training set is highly imbalanced, and  classification algorithms tend to classify negative training points better than the positive points. To address this issue, we penalize misclassifying positive training points more by using a larger regularization parameter $r^+$, in contrast to that of negative training points $r^-$ \added[]{(see the loss functions \eqref{eq:svm}, \eqref{eq:lr} and \eqref{eq:gcn} of the classification algorithms in Appendix~\ref{Subsection: ml algorithms}).} In our experiments, we  set $r^- = 1$ and $r^+ = n_{-1}/n_1$, where $n_{-1}$ and $n_1$ are the number of negative and positive points in our training set.

In the testing phase, we can apply the trained model to predict a scalar $z_{i,j}$ for each edge $e_{i,j}$ in an unseen orienteering problem instance, where $i,j = 1, \ldots, n$ and $i\ne j$. For GCN, $z_{i,j}$ is the output of the last layer. For SVM and LR, $z_{i,j}$ is computed as  $z_{i,j} = \bm{w}_*^T\bm{f}^{i,j}+b_*$, where ($\bm{w}_*$, $b_*$) are the optimized parameters, and $\bm{f}^{i,j}$ is the feature vector of edge $e_{i,j}$.  We then feed the predicted value $z_{i,j}$ into the logistic function to normalize it to a range of $[0,1]$:
\begin{equation}
p_{i,j} = \frac{1}{1+e^{-z_{i,j}}}. 
\end{equation}
The value of $p_{i,j}$ approaches 1 if  $z_{i,j}$ approaches infinity; and $p_{i,j}$ approaches 0 when $z_{i,j}$ approaches negative infinity. In this sense, $p_{i,j}$ can be interpreted as the probability of edge $e_{i,j}$ belonging to an optimal solution. In the next subsection, we will explore multiple ways of incorporating the predicted probability values $p_{i,j}$ into ACO to guide its sampling process.

\subsection{Incorporating solution prediction into ACO}\label{Subsection: Incorporating Solution Prediction into ACO}
Recall that the probabilistic model of ACO heavily depends on the \emph{heuristic weight matrix} $\bm{\eta}$, as shown in Eq. (\ref{Eq: probabilistic model}). The $\eta_{i,j}$ value is a `quality' measure of edge $e_{i,j}$, indicating if it is beneficial to include edge $e_{i,j}$ in a solution in order to obtain a large objective value. The $\bm{\eta}$ values are usually set based on a heuristic rule, for instance in the orienteering problem we can set $\eta_{i,j} = s_{j}/c_{i,j}$, where $s_{j}$ is the score of vertex $v_j$ and $c_{i,j}$ is the cost of edge $e_{i,j}$. Here, we use the probabilities ($\bm{p}$) predicted by our ML model to set $\bm{\eta}$ values: $\eta_{i,j} = p_{i,j}$, and compare it  against the heuristic rule: $\eta_{i,j} = s_{j}/c_{i,j}$. We also explore a hybrid approach that sets the $\bm{\eta}$ values to the product of our ML prediction and the heuristic rule: $\eta_{i,j} = p_{i.j} \cdot s_{j}/c_{i,j}$, for each pair of $i,j=1,\ldots,n$ and $i\ne j$. 


The \emph{pheromone matrix} $\bm{\tau}$ is another important parameter of ACO. The $\bm{\tau}$ values are usually initialized uniformly, and are evolved in each iteration of ACO. Instead, we initialize the $\bm{\tau}$ values by our predicted probabilities, i.e., $\tau_{i,j} = p_{i,j}$. By doing this, better $\bm{\tau}$ values hopefully can be evolved more quickly, and thus high-quality solutions can be constructed earlier. As the pheromone values of the MMAS algorithm are restricted to a range of $[\tau_\mathrm{min}, \tau_\mathrm{max}]$, we re-scale  the predicted probabilities $\bm{p}$ to $[\tau_\mathrm{min}, \tau_\mathrm{max}]$. In addition, if the \emph{pheromone trail smoothing} mechanism is triggered, we re-initialize the $\bm{\tau}$ values to the rescaled  probabilities. 

To summarize, we consider three different ways of incorporating our solution prediction into ACO: 
\begin{enumerate}
	\item Set the $\eta_{i,j}$ value to the predicted probability  value $\eta_{i,j} = p_{i,j}$, and initialize $\tau_{i,j}$ uniformly;
	\item Set the $\eta_{i,j}$ value to the product of the predicted probability and a heuristic rule: $\eta_{i,j} = p_{i,j} \cdot s_{j}/c_{i,j}$, and initialize $\tau_{i,j}$ uniformly;
	\item \added[]{Set the $\eta_{i,j}$ value by the heuristic rule: $\eta_{i,j} = s_{j}/c_{i,j}$, and initialize $\tau_{i,j}$ based on the predicted probability value: $\tau_{i,j} = p_{i,j}$.}
\end{enumerate}
\added[]{In our experiments, we will compare our ML-enhanced ACO with the classic ACO that sets the $\eta_{i,j}$ value by the heuristic rule ($\eta_{i,j} = s_{j}/c_{i,j}$) and initializes $\tau_{i,j}$ uniformly.}

\section{Experiments}\label{Section: Experiments}
We empirically show the efficacy of our ML models for enhancing the performance of ACO via solution prediction for solving the orienteering problem. Specifically, we explore different ways of integrating ML prediction and ACO in Section \ref{Subsection: Different Types of Integration} and compare the effects of using different ML algorithms for solution prediction in Section \ref{Subsection: Comparison of Machine Learning Algorithms}. We then test the generalization capability of our model to large synthetic, benchmark, and real-world problem instances in Section \ref{Subsection: Generalization to Larger Problem Instances}, \ref{Subsection: Generalization to Benchmark Problem Instances}, and \ref{Subsection: Generalization to Real-world Problem Instances},  respectively. \added[]{Finally, we compare our method against the state-of-the-art algorithms in Section~\ref{subsec::comparision with State-of-the-art}.} 
  
Our source codes  are written in C++, which will be made publicly available online when the paper gets published. For the ML algorithms, we use the SVM model implemented in the LIBSVM library \citep{chang2011libsvm}, and the LR model implemented in the LIBLINEAR library \citep{fan2008liblinear}. For GCN, we implement it using TensorFlow \added[]{\citep{tensorflow2015-whitepaper}}. Our experiments are conducted on a high performance computing server at Monash University -- MonARCH, using a NVIDIA Tesla P100 GPU and multiple types of CPUs that are at least 2.40GHz. \added[]{Each CPU is equipped with 4GB memory.} 

To construct a training set, we generate $100$ orienteering problem instances with $50$ vertices.  For each vertex, we randomly generate a pair of real numbers between $0$ and $100$ as its coordinates in the Euclidean space. We assign a score of $0$ to the starting and ending vertices, and generate a random integer between $0$ and $100$ as the score for each of the other vertices. The total distance budget (or time budget) is set to an integer randomly generated between $100$ and $400$ for each problem instance. We then use CPLEX to solve these 100 problem instances, among which 90 are solved to optimality within a cutoff time $10,000$ seconds given to each instance. The total time taken to solve the $90$ problem instances to optimality is about $11.5$ hours if using a single CPU, and the time can be significantly reduced if using multiple CPUs. To train the `deep' GCN model, we construct a large-sized training set using all the $90$ solved problem instances which contains $220,500$ training points. To train the `shallow' LR and SVM models, we only use the first $18$ solved problem instances, \added[]{as using more training data cannot further improve the performance of these models.}

\subsection{Efficacy of integrating machine learning into ACO}\label{Subsection: Different Types of Integration}

We investigate whether the performance of ACO can be improved by solution prediction. To do so, we train a linear SVM model on our training set, that takes about $31$ seconds. For testing, we generate 100 problem instances, each with 100 vertices, in the same way as we generate the training instances. We use the trained SVM model to predict a probability $p_{i,j}$ for each edge $e_{i,j}$ in a test problem instance. The prediction time is about $0.7$ second, which is negligible. 
 
We explore three different ways of incorporating the solution prediction into the probabilistic model of ACO, as shown in Section \ref{Subsection: Incorporating Solution Prediction into ACO}. We denote these hybrid models as 1) SVM-ACO$_{\eta}$, that sets  $\eta_{i,j} = p_{i,j}$; 2) SVM-ACO$_{\hat{\eta}}$ that sets  $\eta_{i,j} = p_{i,j} \cdot  s_{j}/c_{i,j}$; and 3) SVM-ACO$_{\tau}$ that initializes $\tau_{i,j}$ based on $p_{i,j}$. We test two ACO variants, AS and MMAS, which are detailed in Section \ref{Subsection: Ant Colony Optimization}. \added[]{The default parameter settings for AS and MMAS are: $\alpha = 1$, $\beta = 1$, $\rho = 0.05$, $\delta = 0.5$, $T_\mathrm{pts} = 100$, and $C=100y^\mathrm{best}$, where $y^\mathrm{best}$ is the best objective value found so far. The values for $\tau_\mathrm{max}$ and $\tau_\mathrm{min}$ are computed based on Eq. (\ref{Eq: tmax}) and (\ref{Eq: tmin}). For MMAS, the iteration-best solution is used to update the pheromone matrix. These parameter values are selected based on the original papers \citep{dorigo1996ant,stutzle2000max} and our preliminary experimental study.} 

\begin{figure*}[!t]
	\centering
	\includegraphics[scale=1]{ 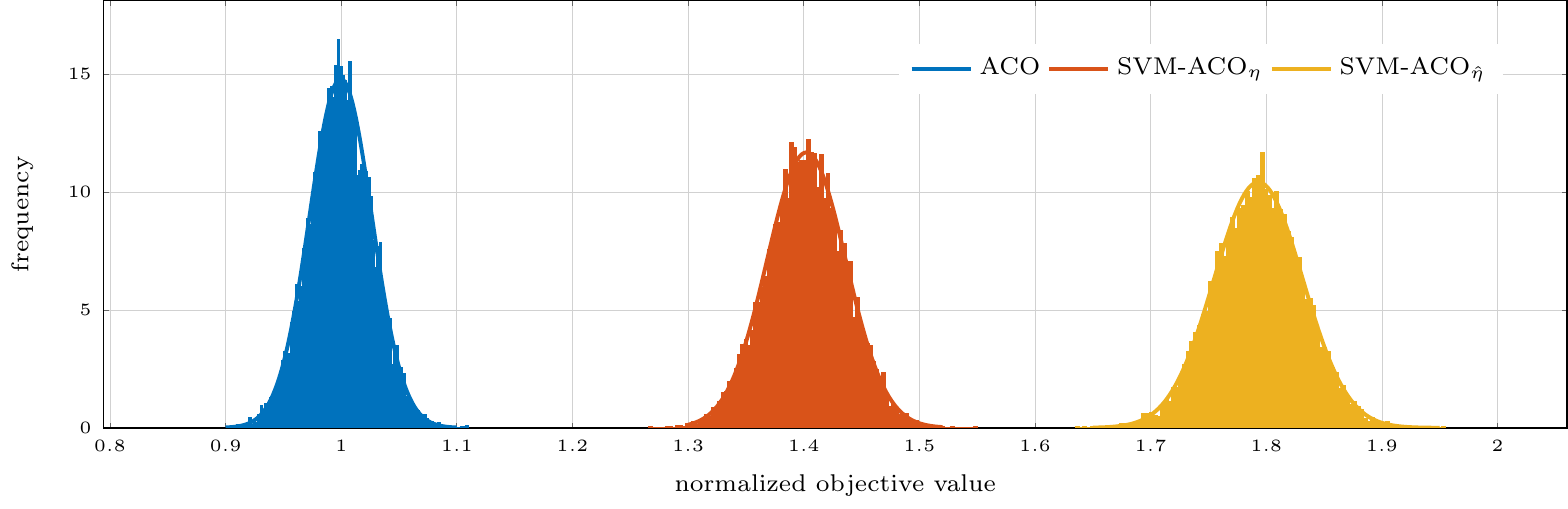}
	\caption{The distribution of the objective values generated by the ACO, SVM-ACO$_{\eta}$, and SVM-ACO$_{\hat{\eta}}$ algorithms in the first iteration when tested on the orienteering problems of size $100$. The objective values  are normalized by the mean objective value generated by ACO. }
	\label{Figrue: The distribution of objective values}
\end{figure*}

To show the efficacy of our ML prediction, we first compare the initial probabilistic models of SVM-ACO$_{\eta}$ and SVM-ACO$_{\hat{\eta}}$ against that of the classic ACO algorithm without ML enhancement. Note that the initial probabilistic model of SVM-ACO$_{\tau}$ is the same as that of SVM-ACO$_{\hat{\eta}}$ under our parameter settings. Moreover, the initial probabilistic models of the two ACO variants, AS and MMAS are also identical. We use the initial probabilistic models to sample 10,000 solutions for each test problem instance, and plot the distribution of averaged normalized objective values in Figure~\ref{Figrue: The distribution of objective values}. The objective values of the sample solutions are normalized by the mean objective value generated by the classic ACO algorithm. The normalized objective values are then averaged across 100 test problem instances. We can observe that the average objective values generated by SVM-ACO$_\eta$ is about 40\% better than that of the classic ACO algorithm without ML enhancement. The only difference between these two algorithms is that SVM-ACO$_\eta$ sets $\eta_{i,j}$ based on predicted probability $p_{i,j}$, while ACO sets $\eta_{i,j}$ based on a heuristic rule $s_{j}/c_{i,j}$. In this sense, our ML prediction is more `greedy' than the heuristic rule.  Furthermore, by setting $\eta_{i,j}$ to the product of our predicted probability and the heuristic rule, the resulted algorithm SVM-ACO$_{\hat{\eta}}$ improves over  ACO  by  80\% in terms of the objective values generated in the first iteration.

\begin{figure*}[!t]
	\centering
	\includegraphics[scale=1]{ 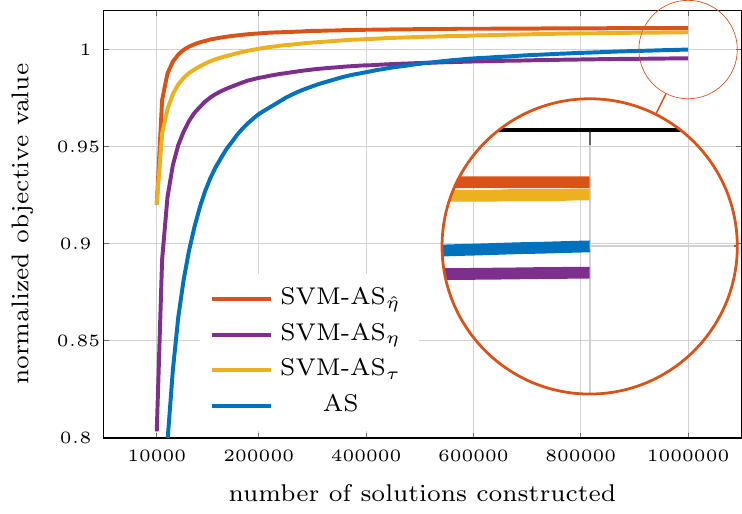}
	\hspace{0.5cm}
	\includegraphics[scale=1]{ 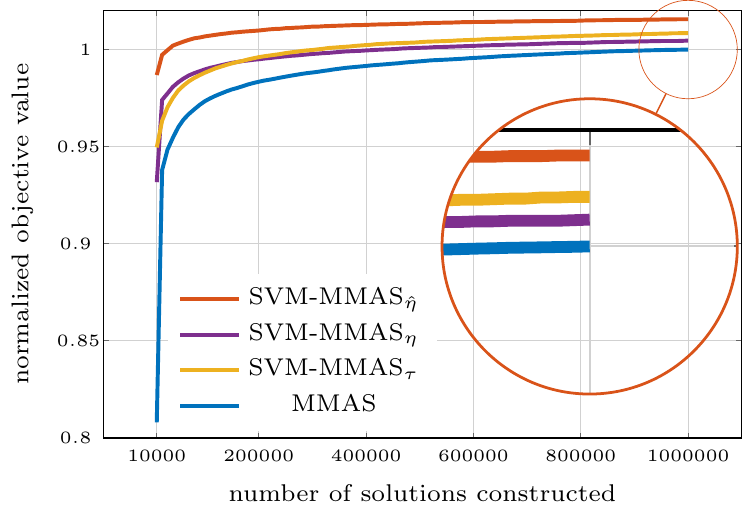}
	\caption{The convergence curves of the ACO (i.e., AS or MMAS),  SVM-ACO$_\tau$, SVM-ACO$_\eta$ and SVM-ACO$_{\hat{\eta}}$ algorithms when used to solve the orienteering problems of size $100$. The objective values are normalized by the best objective value found by ACO and are averaged across 100 instances.}
	\label{Figure: The convergence curves of the Ant System (AS) and Max-Min Ant System (MMAS) with different types of enhancements}
\end{figure*}

We then compare the ACO algorithms (AS or MMAS) enhanced by ML prediction against the classic ACO algorithms, when solving the test problem instances. The number of solutions to be constructed is set to $10000n$ for each algorithm, where $n$ is problem dimensionality. The population size of AS is set to $100n$ and that of MMAS is $n$, because MMAS only uses a single best solution to update the pheromone matrix in each iteration, and thus it benefits more from relatively small population size and more iterations. The objective values generated by each algorithm are normalized by the best objective value found by AS (or MMAS), and are averaged across 100 problem instances and 25 independent runs. The curves of normalized objective value v.s. number of solutions constructed is shown in Figure~\ref{Figure: The convergence curves of the Ant System (AS) and Max-Min Ant System (MMAS) with different types of enhancements}. These convergence curves can show not only the final objective values generated by the algorithms but also their converging speed. We can observe that the performances of both AS and MMAS in finding high-quality solutions are greatly enhanced by ML prediction. Significantly, the solution generated by SVM-MMAS$_{\hat{\eta}}$ at 4\% of computational budget is already better than the final solution produced by MMAS.  Furthermore, the enhanced AS and MMAS algorithms are generally able to find a better solution at the end of a run, except for the SVM-AS$_{\eta}$ algorithm that may have an issue of premature convergence. We note that the hybridization SVM-ACO$_{\hat{\eta}}$  works the best; it improves over the classic ACO by more than 1\% in terms of the final solution quality generated. This improvement is larger if less computational budget is allowed. \added[]{Hence, we will only employ the hybridization ($\hat{\eta}$) that sets $\eta_{i,j} = p_{i,j} \cdot  s_{j}/c_{i,j}$ in the rest of the paper.}

\subsection{Sensitivity to machine learning algorithms}\label{Subsection: Comparison of Machine Learning Algorithms}

We take the SVM-ACO algorithm and replace SVM by LR and GCN to see if the performance of our hybrid algorithm is sensitive to the ML algorithm used in training. We train a separate model with LR and GCN on our training set. For GCN, we use 20 layers and each hidden layer has 32 neurons. The learning rate is set to $0.001$ and the number of epochs is $100$. The training time for LR is about $27$ seconds and that for GCN is about $1000$ seconds. 

\begin{figure*}[!t]
	\centering
	\includegraphics[scale=1]{ 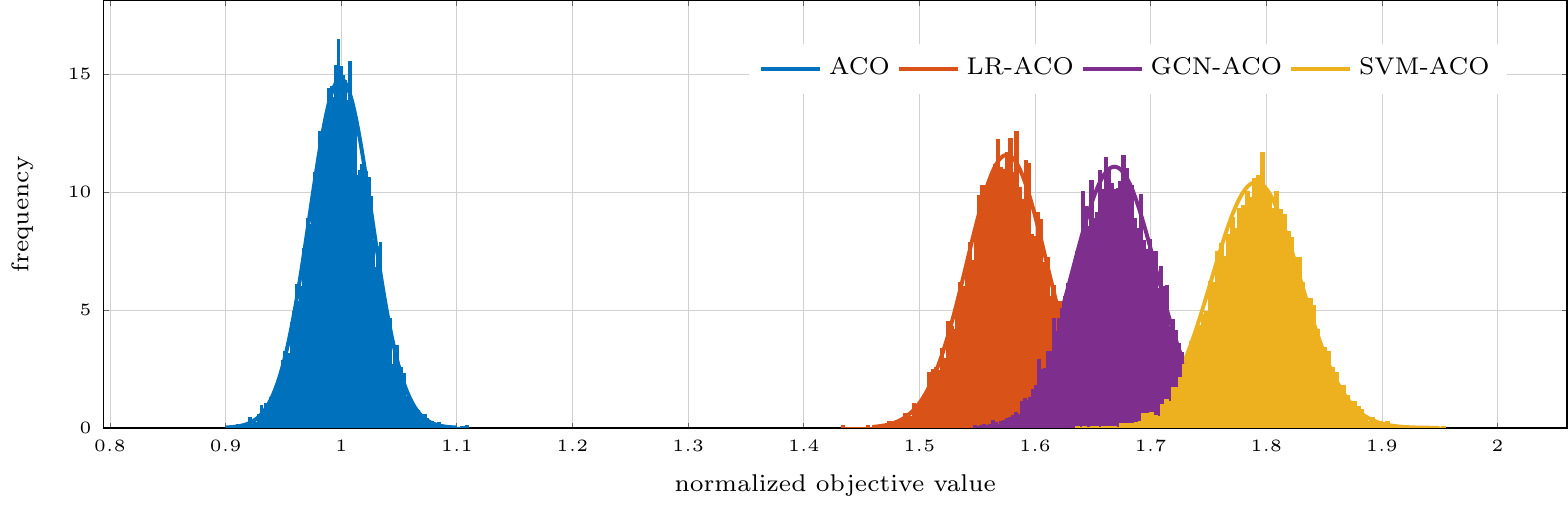}
	\caption{The distribution of objective values generated by the ACO,  SVM-ACO, GCN-ACO, and LR-ACO algorithms in the first iteration when tested on the orienteering problems of size $100$. The objective values are normalized by the mean objective value generated by ACO.}
	\label{Figure: The distribution of objective values generated by AS (or MMAS) in the first iteration when hybridized with different classifiers to solve the orienteering problems of size $100$.}
\end{figure*}

Similar as before, we compare the initial probabilistic models of the SVM-ACO, LR-ACO, GCN-ACO and ACO algorithms, and plot the distribution of  objective values generated by each probabilistic model in Figure~\ref{Figure: The distribution of objective values generated by AS (or MMAS) in the first iteration when hybridized with different classifiers to solve the orienteering problems of size $100$.}. We can observe that no matter which one of the three learning algorithms is used, the ACO enhanced by solution prediction significantly improves over the classic ACO by more than 50\% in terms of the objective values generated in the first iteration. Among the three learning algorithms, the LR performs the worst and SVM is the best. This is a bit surprising as the simple linear SVM model performs slightly better than the  deep GCN model in this context. Note that we have not done any fine-tuning for the GCN model, and we suspect that the performance of GCN  may be further improved by tuning hyper-parameters. However, a thorough evaluation along this line \added[]{requires significantly more computational resources} and is beyond the scope of this paper. 

\begin{figure*}[!t]
	\centering
	\includegraphics[scale=1]{ 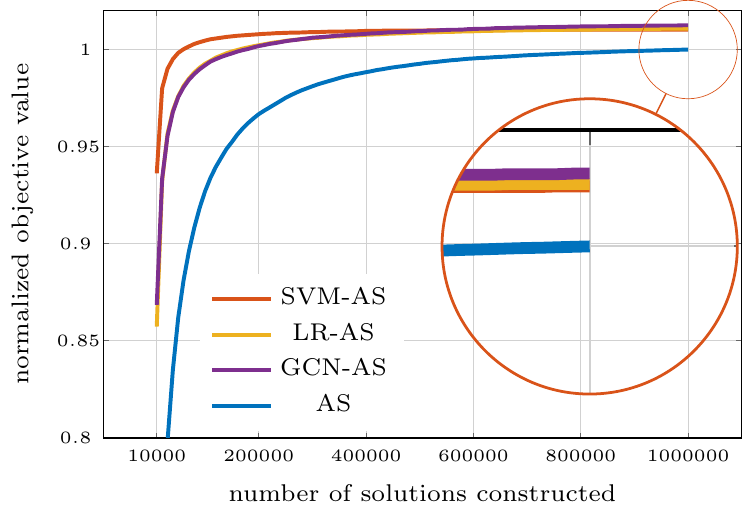}
	\hspace{0.5cm}
	\includegraphics[scale=1]{ 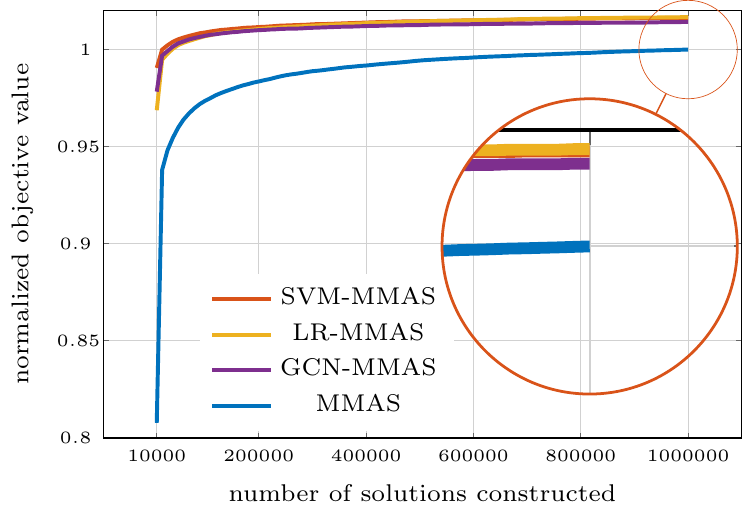}
	\caption{The convergence curves of the ACO (i.e., AS or MMAS), SVM-ACO, GCN-ACO, and  LR-ACO algorithms when used to solve the orienteering problems of size $100$. The objective values are normalized by the best objective value found by ACO and are averaged across 100 instances.}
	\label{Figure: The convergence curves of the Ant System (AS) and Max-Min Ant System (MMAS) when hybridized with different classifiers to solve the orienteering problems of size $100$.}
\end{figure*}

We also compare the performance of the four algorithms for solving the test problem instances, and the averaged convergence curves are shown in Figure~\ref{Figure: The convergence curves of the Ant System (AS) and Max-Min Ant System (MMAS) when hybridized with different classifiers to solve the orienteering problems of size $100$.}.  The results show that the ACO algorithms (i.e., AS or MMAS) enhanced by different ML predictions consistently outperform the classic ACO in finding high-quality solutions. Whilst the initial objective values found by SVM-ACO, LR-ACO, and GCN-ACO  are different, the final solutions generated by these algorithms are of a similar quality after many iterations of ACO sampling. \added[]{In the rest of this paper, we will only use SVM as our ML model.}

\subsection{Generalization to larger problem instances}\label{Subsection: Generalization to Larger Problem Instances}

We test the generalization of our SVM-ACO algorithm to larger orienteering problem instances. To do so, we randomly generate larger problem instances with dimensionality 200, 300, 400 and 500, each with 100 problem instances, for testing. We then apply the SVM-ACO model, which is trained on small  problem instances of dimensionality 50, to solve each of the larger test problem instance, compared against the classic ACO. The parameter settings for the two ACO variants, AS and MMAS are the same as before.

\begin{table*}[!t]
	\centering
 	\small
	\caption{The best objective values obtained by the AS, SVM-AS, MMAS and SVM-MMAS algorithms averaged across 100 problem instances in each problem set. A series of \added[]{Wilcoxon signed-rank tests} are performed between each pair of algorithms (i.e., AS vs SVM-AS and MMAS vs SVM-MMAS), and the p-values are reported. The statistically significantly better results are highlighted in bold (p-value $<$ 0.05).}
	\label{table: statistical tests}
	\begin{tabular}{@{\extracolsep{6pt}}lrrrrrr@{}}
		\toprule
		Dataset  & AS & SVM-AS & p-value  & MMAS & SVM-MMAS  & p-value \\ \cline{1-1}   \cline{2-4}  \cline{5-7} 
		size 200 & 3043.22 & \bf{3102.74} & 3.02e-12 & 3068.50 & 	\bf{3254.87} & 3.39e-15 \\
		size 300 & 3674.04 & \bf{3739.04} & 3.42e-12 & 3637.37 & 	\bf{3944.72} & 8.86e-17\\
		size 400 & 4278.65 & \bf{4392.26} & 4.05e-15 & 4207.71 &	\bf{4665.61} & 4.01e-18 \\
		size 500 & 4752.65 & \bf{4866.93} & 6.10e-14 & 4612.65 &	\bf{5167.52} & 1.05e-17 \\\bottomrule
	\end{tabular}
\end{table*}
 \begin{figure*}[!t]
	\centering
	\subfloat[\scriptsize size 200]{
		\includegraphics[scale=1]{ 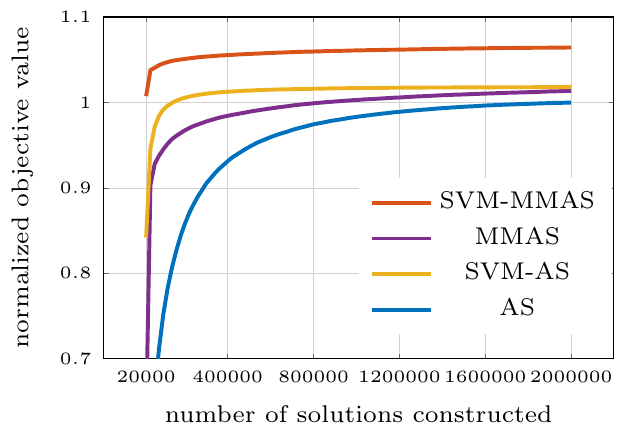}
	}\hspace{0.5cm}	
	\subfloat[\scriptsize size 300]{
		\includegraphics[scale=1]{ 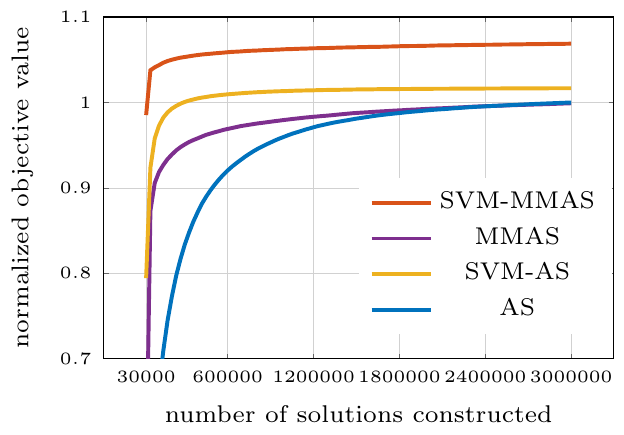}
	}
	
	\subfloat[\scriptsize size 400]{
		\includegraphics[scale=1]{ 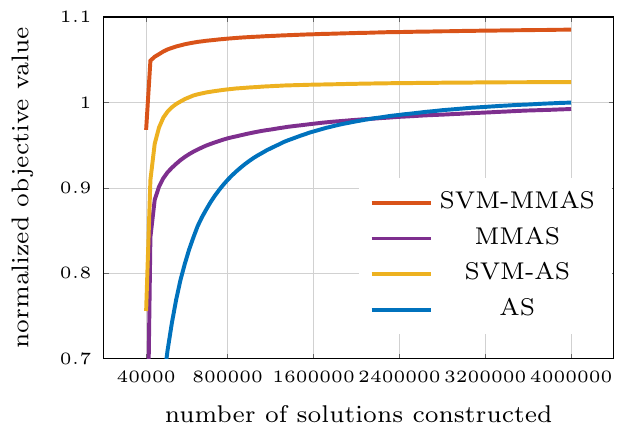}
	}\hspace{0.5cm}
	\subfloat[\scriptsize size 500]{
		\includegraphics[scale=1]{ 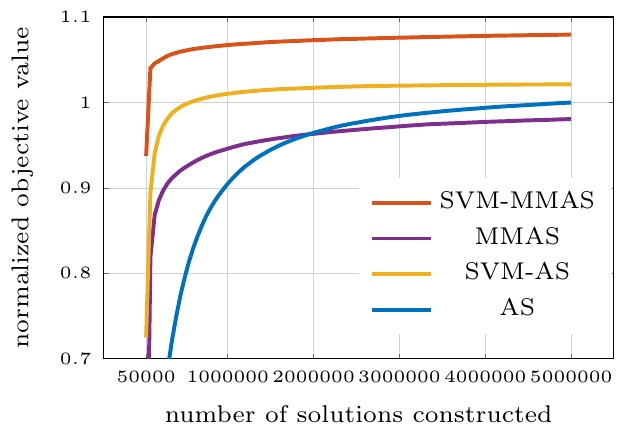}
	}
	\caption{The convergence curves of the AS, SVM-AS, MMAS and SVM-MMAS algorithms, when used to solve the larger orienteering problem instances. The objective values are normalized by the best objective value found by AS and are averaged across 100 instances.} 
	\label{Figure: Convergence curves on larger problems.}
\end{figure*}

The best objective values obtained by each algorithm averaged across 100 problem instances for each problem set are presented in Table~\ref{table: statistical tests}, and the averaged convergence curves  are shown in Figure~\ref{Figure: Convergence curves on larger problems.}. First, we can observe that our ML model trained on small problem instances generalizes very well to larger test problem instances, in the sense that it consistently boosts the performance of both AS and MMAS when solving the larger problem instances. Furthermore, this improvement  becomes more significant as the problem dimensionality increases from 200 to 500. The SVM-MMAS algorithm is clearly the best performing one among the four algorithms tested. Significantly, SVM-MMAS improves over MMAS by more than 10\% in terms of the best objective values generated for the problem instances of dimensionality 500.

\subsection{Generalization to benchmark problem instances} \label{Subsection: Generalization to Benchmark Problem Instances}

We evaluate the generalization capability of our SVM-ACO algorithm on a set of benchmark instances used in \citep{chao1996fast}. Each instance has 66 vertices, the locations of which form a square shape. The distance budget is varied from 5 to 130 in increments of 5, resulting in 26 instances in total. We apply the SVM-ACO algorithm, trained on randomly generated problem instances, to solve the square-shaped benchmark problem instances, compared against the classic ACO under the same parameter settings. 

\begin{figure}[!t]
	\centering
	\includegraphics[scale=0.9]{ 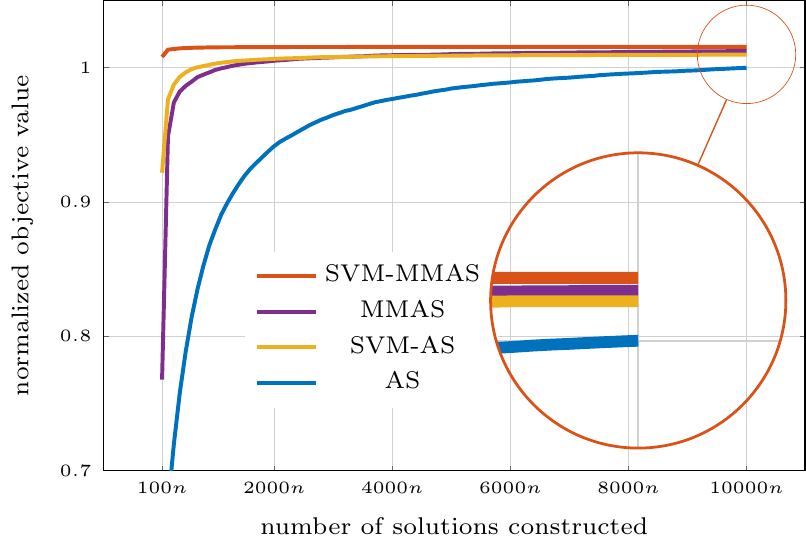}
	\caption{The convergence curves of the AS, SVM-AS, MMAS and SVM-MMAS  algorithms, when used to solve the benchmark problem instances. The objective values are normalized by the best objective value found by AS and are averaged across 26 instances.} 
	\label{Figure: Convergence curves on benchmark problem instances.}
\end{figure}

The averaged convergence curves of the algorithms when used to solve the benchmark problem instances are shown in Figure~\ref{Figure: Convergence curves on benchmark problem instances.}. The results show that our ML model trained on randomly generated instances generalizes well to the square-shaped benchmark instances. In particular, the SVM-MMAS algorithm is able to find a high-quality solution at the very early stage of the search process, comparing to MMAS. The mean and standard deviation of the best objective values generated by each algorithm across 25 runs for each problem instance are presented in Table~\ref{Table: The optimization results of AS, SVM-AS, MMAS and SVM-MMAS on benchmark instances}. The statistically significantly better results obtained by the SVM-ACO and ACO algorithms are highlighted in bold, based on the \added[]{Wilcoxon signed-rank tests} with a significance level of 0.05. We can observe that on easy instances in which the distance budget is small, both the SVM-ACO and ACO algorithms are able to find the optimal solutions by the end of a run consistently. However, on hard instances in which the distance budget is large, the solution quality generated by  SVM-ACO is statistically significantly better than that by the classic ACO algorithm. 

\begin{table*}[!t]
	\centering
 	\small
	\caption{The best objective values generated by the AS, SVM-AS, MMAS and SVM-MMAS algorithms on the benchmark problem instances. The statistically significantly better results generated by SVM-AS as opposed to AS (and SVM-MMAS as opposed to MMAS) are highlighted in bold, according to \added[]{Wilcoxon signed-rank tests} with a significance level of 0.05.}
	\label{Table: The optimization results of AS, SVM-AS, MMAS and SVM-MMAS on benchmark instances}
	\begin{tabular}{@{\extracolsep{6pt}}llrrrrrrrr@{}}
		\toprule
		\multirow{2}{*}{Datasets} & \multirow{2}{*}{Budget} & \multicolumn{2}{c}{AS}  &  \multicolumn{2}{c}{SVM-AS}  & \multicolumn{2}{c}{MMAS}  & \multicolumn{2}{c}{SVM-MMAS}  \\
		\cline{3-6}  \cline{7-10} 
		& & mean & std & mean & std & mean & std & mean & std \\ \cline{1-1}  \cline{2-2}  \cline{3-4}  \cline{5-6}  \cline{7-8}  \cline{9-10} 
		set\_66\_1\_005 & 5 & 10.00   & 0.00  & 10.00   & 0.00  & 10.00   & 0.00 & 10.00   & 0.00 \\
		set\_66\_1\_010 & 10 & 40.00   & 0.00  & 40.00   & 0.00  & 40.00   & 0.00 & 40.00   & 0.00 \\
		set\_66\_1\_015 & 15 & 120.00  & 0.00  & 120.00  & 0.00  & 120.00  & 0.00 & 120.00  & 0.00 \\
		set\_66\_1\_020 & 20 & 205.00  & 0.00  & 205.00  & 0.00  & 205.00  & 0.00 & 205.00  & 0.00 \\
		set\_66\_1\_025 & 25 & 290.00  & 0.00  & 290.00  & 0.00  & 290.00  & 0.00 & 290.00  & 0.00 \\
		set\_66\_1\_030 & 30 & 400.00  & 0.00  & 400.00  & 0.00  & 400.00  & 0.00 & 400.00  & 0.00 \\
		set\_66\_1\_035 & 35 & 465.00  & 0.00  & 465.00  & 0.00  & 465.00  & 0.00 & 465.00  & 0.00 \\
		set\_66\_1\_040 & 40 & 575.00  & 0.00  & 575.00  & 0.00  & 575.00  & 0.00 & 575.00  & 0.00 \\
		set\_66\_1\_045 & 45 & 647.60  & 2.55  & \bf{649.20}  & 1.87  & 648.60  & 2.29 & \bf{650.00}  & 0.00 \\
		set\_66\_1\_050 & 50 & 730.00  & 0.00  & 730.00  & 0.00  & 729.20  & 3.12 & 730.00  & 0.00 \\
		set\_66\_1\_055 & 55 & 820.40  & 2.00  & \bf{823.60}  & 3.39  & 823.00  & 2.50 & \bf{824.80}  & 1.00 \\
		set\_66\_1\_060 & 60 & 904.80  & 6.84  & \bf{910.20}  & 5.68  & 914.20  & 2.36 & 915.00  & 0.00 \\
		set\_66\_1\_065 & 65 & 972.40  & 5.02  & \bf{976.60}  & 5.72  & 980.00  & 0.00 & 980.00  & 0.00 \\
		set\_66\_1\_070 & 70 & 1055.00 & 8.42  & \bf{1065.20} & 8.72  & 1069.60 & 1.38 & 1070.00 & 0.00 \\
		set\_66\_1\_075 & 75 & 1126.00 & 9.46  & \bf{1138.60} & 4.45  & 1140.00 & 0.00 & 1140.00 & 0.00 \\
		set\_66\_1\_080 & 80 & 1188.20 & 10.09 & \bf{1206.80} & 8.52  & 1209.20 & 6.24 & \bf{1215.00} & 0.00 \\
		set\_66\_1\_085 & 85 & 1242.80 & 6.47  & \bf{1265.20} & 6.84  & 1264.60 & 3.20 & \bf{1270.00} & 0.00 \\
		set\_66\_1\_090 & 90 & 1302.60 & 9.03  & \bf{1322.80} & 10.91 & 1329.80 & 6.84 & \bf{1340.00} & 0.00 \\
		set\_66\_1\_095 & 95 & 1354.20 & 10.58 & \bf{1379.40} & 10.24 & 1386.60 & 6.25 & \bf{1394.80} & 1.00 \\
		set\_66\_1\_100 & 100 & 1405.60 & 8.33  & \bf{1436.40} & 11.23 & 1447.20 & 9.36 & \bf{1464.60} & 2.00 \\
		set\_66\_1\_105 & 105 & 1452.00 & 11.64 & \bf{1484.40} & 10.64 & 1508.20 & 8.65 & \bf{1519.20} & 2.77 \\
		set\_66\_1\_110 & 110 & 1496.60 & 10.48 & \bf{1529.60} & 9.46  & 1549.40 & 7.82 & \bf{1560.00} & 0.00 \\
		set\_66\_1\_115 & 115 & 1544.00 & 10.80 & \bf{1576.80} & 9.45  & 1583.80 & 8.07 & \bf{1594.80} & 1.00 \\
		set\_66\_1\_120 & 120 & 1582.40 & 8.55  & \bf{1617.40} & 9.03  & 1623.00 & 8.29 & \bf{1634.60} & 2.00 \\
		set\_66\_1\_125 & 125 & 1615.40 & 6.91  & \bf{1654.60} & 9.46  & 1654.60 & 5.94 & \bf{1670.00} & 0.00 \\
		set\_66\_1\_130 & 130 & 1649.40 & 5.83  & \bf{1675.80} & 4.72  & 1675.00 & 3.82 & \bf{1680.00} & 0.00 \\\bottomrule
	\end{tabular}
\end{table*}

\subsection{Generalization to real-world problem instances}\label{Subsection: Generalization to Real-world Problem Instances}

\begin{figure}[!t]
	\centering
	\includegraphics[scale=0.9]{ 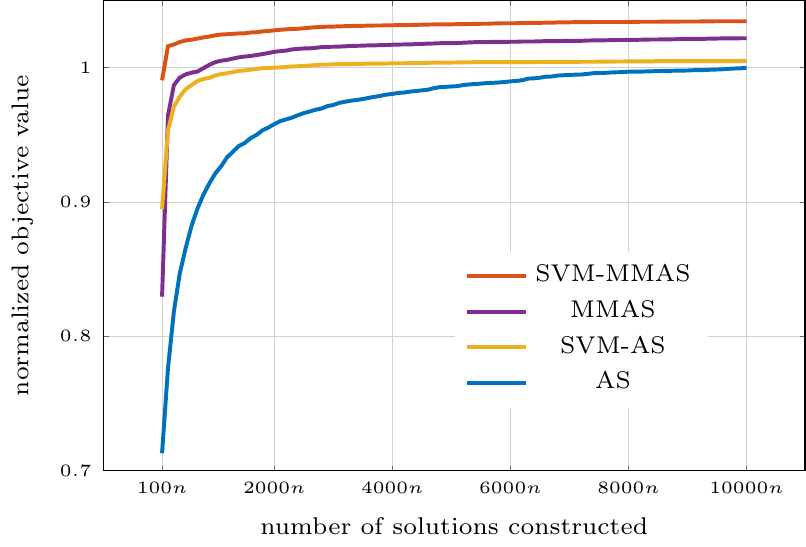}
	\caption{The convergence curves of the AS, SVM-AS, MMAS and SVM-MMAS algorithms, when used to solve the  real-world problem instances. The objective values are normalized by the best objective value found by AS and are averaged across six instances.}
	\label{Figure: Convergence curves on real-world problems.}
\end{figure}

\begin{table*}[!t]
	\centering
 	\small
	\caption{
		The best objective values generated by the AS, SVM-AS, MMAS and SVM-MMAS algorithms on the real-world problem instances. The statistically significantly better results generated by  SVM-AS as opposed to AS (and SVM-MMAS as opposed to MMAS) are highlighted in bold, according to \added[]{Wilcoxon signed-rank tests} with a significance level of 0.05.} 
	\label{Table: The optimization results of AS, SVM-AS, MMAS and SVM-MMAS}
	\begin{tabular}{@{\extracolsep{6pt}}llrrrrrrrr@{}}
		\toprule
		\multirow{2}{*}{Datasets} & \multirow{2}{*}{Dimension} & \multicolumn{2}{c}{AS}  &  \multicolumn{2}{c}{SVM-AS}  & \multicolumn{2}{c}{MMAS}  & \multicolumn{2}{c}{SVM-MMAS}  \\
		\cline{3-6}  \cline{7-10} 
		& & mean & std & mean & std & mean & std & mean & std \\ \cline{1-1}  \cline{2-2}  \cline{3-4}  \cline{5-6}  \cline{7-8}  \cline{9-10} 
		Berlin & 97 & 188.39 & 1.08 & \bf{191.00} & 1.70 & 192.63 & 0.93 & \bf{195.46} & 0.64 \\
		Copenhagen & 81 & 227.28 & 1.90 & 228.01 & 2.14 & 232.21 & 1.61 & \bf{235.62} & 1.79 \\
		Istanbul & 154 & 206.11 & 1.68 & \bf{208.36} & 1.79 & 209.11 & 1.25 & \bf{215.12} & 1.25 \\
		London & 114 & 172.88 & 0.92 & 172.52 & 0.70 & 175.82 & 0.64 & 176.02 & 1.52 \\
		Paris & 117 & 153.81 & 0.76 & 153.47 & 2.21 & 158.83 & 1.41 & 159.19 & 1.78 \\
		Prague & 78 & 242.73 & 1.34 & \bf{244.52} & 1.62 & 248.48 & 1.26 & \bf{251.73} & 1.20\\\bottomrule
	\end{tabular}
\end{table*}

We further test the generalization of our SVM-ACO algorithm to real-world problem  instances -- tourist trip planning, where the goal is to plan an itinerary that visits a subset of attractions in a city within a given distance (or time) budget such that the total collected `popularity score' is maximized. We use six datasets published in a trajectory-driven tourist trip planning system \citep{wang2018trip} to test our model. Each dataset corresponds to a city in Europe. The starting and ending vertex of a dataset is a hotel randomly chosen from the  corresponding city, and the other vertices are attractions for visiting. The coordinates of each vertex are its geographic location: latitude and longitude; and the distance between two vertices is the geographical distance between them.  The popularity score of an attraction is calculated based on how many people have visited that attraction: $s_i = \log_2(n_i + 1)$, where $n_i$ is the number of trajectories that have been to that attraction. The trajectory data was originally crawled from the Triphobo website by Wang \emph{et al.} \citep{wang2018trip}. The dimensionality of these six datasets varies from 81 to 154. The total distance budget is set to 50 kilometers for each dataset. We take the SVM-ACO model trained on synthetic problem instances and test it on these real-world problem instances.

The average convergence curves of the AS, SVM-AS, MMAS  and SVM-MMAS algorithms when used to solve the real-world problem instances are shown in Figure~\ref{Figure: Convergence curves on real-world problems.}. The results show that our ML model trained on synthetic problem instances generalizes well to real-world problem instances; the ML model  speeds up both AS and MMAS in finding high-quality solutions for the real-world instances. Overall, the SVM-MMAS algorithm achieves the best solution quality, and improves over MMAS by 1.24\% on average. The mean and standard deviation of the best objective values generated by each algorithm across 25 runs for each problem instance are presented in Table~\ref{Table: The optimization results of AS, SVM-AS, MMAS and SVM-MMAS}. We can see that both AS and MMAS consistently find an equally well or statistically significantly better solution when enhanced by our ML prediction.

\subsection{\added[]{Comparison with state-of-the-art algorithms}}
\label{subsec::comparision with State-of-the-art}

\added[]{
We take the best-performing algorithm SVM-MMAS and compare it against three state-of-the-art heuristics, Evolutionary Algorithm for Orienteering Problem (EA4OP) \citep{kobeaga2018efficient}, GRASP with Path Relinking (GRASP-PR) \citep{campos2014grasp} and 2-Parameter Iterative Algorithm (2P-IA) \citep{silberholz2010effective}. These three state-of-the-art algorithms all use effective local search methods. For a fair comparison, we also use a local search method to improve the solutions sampled by SVM-MMAS. Specifically, we use the well-known 2-opt local search method \citep{lin1965computer} to reduce the path length of the best solution constructed in each iteration of SVM-MMAS. Candidate vertices (i.e., those have not been visited) are greedily inserted into the path until a local optimum is found. Note that the 2-opt local search method is repeatedly applied when a candidate vertex is inserted into the path, attempting to reduce the path length. 


We compare the performance of our algorithm (denoted as SVM-MMAS-LS) with the state-of-the-art heuristics on a set of benchmark problem instances \citep{fischetti1998solving}, of which the number of vertices ranges from 48 to 400. These instances (generation 3) were generated based on the TSP library, and are available from the OP library (\url{https://github.com/bcamath-ds/OPLib}). For each problem instance, we run our SVM-MMAS-LS algorithm 10 times, and record the best solution found and the average runtime used, following \citep{kobeaga2018efficient}. The parameters of our algorithm are set as before with four exceptions:}
\begin{enumerate}
    \item \added[]{We use a different termination criterion for our SVM-MMAS-LS algorithm, i.e., if the best solution found cannot be improved for $T_\mathrm{ter}$ consecutive iterations, the algorithm is terminated. We will test two different values for $T_\mathrm{ter}$: 200 and 500.} 
    \item \added[]{We have tested the use of both the iteration-best solution and the global-best solution to update the pheromone matrix, and found that using the global-best solution can generate an optimality gap that is 44\% better than that of using the iteration-best solution on average. Therefore, we will use the global-best solution to update the pheromone matrix.} 
    \item \added[]{We have tested three different population sizes \{50, 100, $n$\} and found that using a larger population size can generate a slightly smaller optimality gap but significantly increases the runtime. Hence, we will set the population size to 50.} 
    \item \added[]{We use the SVM model (primal formulation) implemented in the LIBLINEAR library, which is faster than that of the LIBSVM library (dual formulation) in prediction. Furthermore, we reduce the sample size $m$ to $10n$, to gain computational efficiency.}
\end{enumerate}

\added[]{The results are presented in Table~\ref{tab::state-of-the-art}. Note that the results of EA4OP, GRASP-PR and 2P-IA are taken from Table A.5 of \citep{kobeaga2018efficient}. We can observe that our SVM-MMAS-LS algorithm with $T_\mathrm{ter}=200$ is already competitive with the three state-of-the-art heuristics. On average, SVM-MMAS-LS achieves a smaller optimality gap than the other three algorithms using similar runtime (see the last row of Table~\ref{tab::state-of-the-art}). When $T_\mathrm{ter}$ is increased from 200 to 500, the average optimality gap can be further reduced, but at an expense of longer runtime.}  

\added[]{Finally, we would like to remark that our primary aim is \emph{not} to develop the best algorithm for solving the orienteering problem. Instead, our aim is to boost the performance of ACO via solution prediction and ML. Therefore, our experiments have mainly focused on showing whether the ML-enhanced ACO algorithm outperforms the classic ACO. In fact, our ML-enhanced ACO is not confined to solving the orienteering problem. In Appendix~\ref{sec::mwcp}, we adapt our ML-enhanced ACO algorithm to solve the maximum weighted clique problem and show that it is competitive compared to the state-of-the-art algorithms.} 


\begin{table}[!t]
\caption{\added[]{The comparison between our SVM-MMAS-LS algorithm and three state-of-the-art heuristics on a set of benchmark problem instances. The column `Opt' presents the optimal objective value. For each algorithm, the best objective value found, optimality gap (\%) and the average time used (in second) are presented. The best optimality gap is highlighted in bold.}}
\label{tab::state-of-the-art}
\resizebox{\textwidth}{!}{
\begin{tabular}{@{\extracolsep{6pt}}lrrrrrrrrrrrrrrrr@{}}
\toprule
 \multirow{2}{*}{Instance} & \multirow{2}{*}{Opt}  & \multicolumn{3}{c}{2P-IA}       & \multicolumn{3}{c}{GRASP-PR}  & \multicolumn{3}{c}{EA4OP}  & \multicolumn{3}{c}{SVM-MMAS-LS (200)}      & \multicolumn{3}{c}{SVM-MMAS-LS (500)}  \\\cline{3-5}\cline{6-8}\cline{9-11}\cline{12-14} \cline{15-17}
 &    & best           & gap   & time  & best          & gap  & time  & best  & gap   & time  & best     & gap  & time  & best     & gap  & time  \\\midrule
att48    & 1049  & 1049           & 0.00  & 0.13  & 1049          & 0.00 & 0.18  & 1049  & 0.00  & 0.26  & 1049     & 0.00 & 0.22  & 1049     & 0.00 & 0.39  \\
gr48     & 1480  & 1480           & 0.00  & 0.07  & 1480          & 0.00 & 0.20  & 1480  & 0.00  & 0.13  & 1480     & 0.00 & 0.21  & 1480     & 0.00 & 0.43  \\
hk48     & 1764  & 1764           & 0.00  & 0.09  & 1764          & 0.00 & 0.14  & 1764  & 0.00  & 0.22  & 1764     & 0.00 & 0.23  & 1764     & 0.00 & 0.36  \\
eil51    & 1399  & 1399           & \bf{0.00}  & 0.12  & 1399          & \bf{0.00} & 0.17  & 1398  & 0.07  & 0.22  & 1399     & \bf{0.00} & 0.25  & 1399     & \bf{0.00} & 0.50  \\
berlin52 & 1036  & 1036           & \bf{0.00}  & 0.19  & 1036          & \bf{0.00} & 0.30  & 1034  & 0.19  & 0.64  & 1036     & \bf{0.00} & 0.20  & 1036     & \bf{0.00} & 0.39  \\
brazil58 & 1702  & 1702           & 0.00  & 0.13  & 1702          & 0.00 & 0.33  & 1702  & 0.00  & 0.71  & 1702     & 0.00 & 0.29  & 1702     & 0.00 & 0.48  \\
st70     & 2108  & 2108           & 0.00  & 0.24  & 2108          & 0.00 & 0.37  & 2108  & 0.00  & 0.31  & 2108     & 0.00 & 0.36  & 2108     & 0.00 & 0.79  \\
eil76    & 2467  & 2461           & 0.24  & 0.30  & 2462          & 0.20 & 0.44  & 2467  & \bf{0.00}  & 0.36  & 2462     & 0.20 & 0.51  & 2467     & \bf{0.00} & 1.05  \\
pr76     & 2430  & 2430           & 0.00  & 0.26  & 2430          & 0.00 & 0.56  & 2430  & 0.00  & 0.57  & 2430     & 0.00 & 0.48  & 2430     & 0.00 & 1.12  \\
gr96     & 3170  & 3170           & \bf{0.00}  & 0.39  & 3153          & 0.54 & 1.07  & 3166  & 0.13  & 1.41  & 3170     & \bf{0.00} & 0.93  & 3170     & \bf{0.00} & 1.49  \\
rat99    & 2908  & 2896           & 0.41  & 0.47  & 2881          & 0.93 & 0.80  & 2886  & 0.76  & 0.78  & 2870     & 1.31 & 0.54  & 2908     & \bf{0.00} & 1.32  \\
kroA100  & 3211  & 3211           & \bf{0.00}  & 0.30  & 3211          & \bf{0.00} & 1.16  & 3180  & 0.97  & 0.38  & 3206     & 0.16 & 0.76  & 3211     & \bf{0.00} & 1.69  \\
kroB100  & 2804  & 2804           & \bf{0.00}  & 0.46  & 2804          & \bf{0.00} & 1.34  & 2785  & 0.68  & 0.51  & 2804     & \bf{0.00} & 0.64  & 2804     & \bf{0.00} & 1.29  \\
kroC100  & 3155  & 3155           & \bf{0.00}  & 0.38  & 3149          & 0.19 & 0.86  & 3155  & \bf{0.00}  & 0.44  & 3149     & 0.19 & 0.70  & 3140     & 0.48 & 1.20  \\
kroD100  & 3167  & 3123           & 1.39  & 0.65  & 3167          & \bf{0.00} & 1.18  & 3141  & 0.82  & 0.58  & 3147     & 0.63 & 0.58  & 3151     & 0.51 & 1.55  \\
kroE100  & 3049  & 3027           & 0.72  & 0.56  & 3049          & \bf{0.00} & 1.48  & 3049  & \bf{0.00}  & 0.47  & 3049     & \bf{0.00} & 0.58  & 3049     & \bf{0.00} & 1.15  \\
rd100    & 2926  & 2924           & 0.07  & 0.62  & 2924          & 0.07 & 0.90  & 2923  & 0.10  & 0.48  & 2926     & \bf{0.00} & 0.71  & 2926     & \bf{0.00} & 1.85  \\
eil101   & 3345  & 3333           & 0.36  & 0.46  & 3322          & 0.69 & 0.76  & 3345  & \bf{0.00}  & 0.56  & 3335     & 0.30 & 0.92  & 3322     & 0.69 & 2.02  \\
lin105   & 2986  & 2986           & \bf{0.00}  & 0.54  & 2986          & \bf{0.00} & 1.89  & 2973  & 0.44  & 2.09  & 2986     & \bf{0.00} & 0.68  & 2986     & \bf{0.00} & 1.92  \\
pr107    & 1877  & 1877           & \bf{0.00}  & 0.29  & 1877          & \bf{0.00} & 1.15  & 1802  & 4.00  & 0.82  & 1875     & 0.11 & 0.46  & 1877     & \bf{0.00} & 0.76  \\
gr120    & 3779  & 3736           & 1.14  & 0.96  & 3745          & 0.90 & 1.15  & 3748  & 0.82  & 1.36  & 3687     & 2.43 & 1.23  & 3765     & \bf{0.37} & 3.07  \\
pr124    & 3557  & 3517           & 1.12  & 0.62  & 3549          & \bf{0.22} & 2.41  & 3455  & 2.87  & 0.88  & 3549     & \bf{0.22} & 0.98  & 3549     & \bf{0.22} & 1.63  \\
bier127  & 2365  & 2356           & 0.38  & 1.08  & 2332          & 1.40 & 2.07  & 2361  & \bf{0.17}  & 2.62  & 2336     & 1.23 & 1.57  & 2347     & 0.76 & 3.62  \\
pr136    & 4390  & 4390           & \bf{0.00}  & 0.93  & 4380          & 0.23 & 2.56  & 4390  & \bf{0.00}  & 1.13  & 4312     & 1.78 & 1.31  & 4299     & 2.07 & 2.85  \\
gr137    & 3954  & 3928           & 0.66  & 1.13  & 3926          & 0.71 & 1.89  & 3954  & \bf{0.00}  & 1.88  & 3932     & 0.56 & 1.80  & 3934     & 0.51 & 2.64  \\
pr144    & 3745  & 3633           & 2.99  & 0.77  & 3745          & 0.00 & 3.36  & 3700  & 1.20  & 2.41  & 3745     & \bf{0.00} & 0.96  & 3745     & \bf{0.00} & 2.11  \\
kroA150  & 5039  & 5037           & \bf{0.04}  & 1.26  & 5018          & 0.42 & 3.06  & 5019  & 0.40  & 1.07  & 5011     & 0.56 & 1.56  & 5034     & 0.10 & 3.71  \\
kroB150  & 5314  & 5267           & 0.88  & 1.31  & 5272          & 0.79 & 2.31  & 5314  & \bf{0.00}  & 1.04  & 5177     & 2.58 & 1.68  & 5253     & 1.15 & 4.12  \\
pr152    & 3905  & 3557           & 8.91  & 0.80  & 3905          & \bf{0.00} & 4.07  & 3902  & 0.08  & 3.62  & 3905     & \bf{0.00} & 1.37  & 3905     & \bf{0.00} & 2.85  \\
u159     & 5272  & 5272           & \bf{0.00}  & 1.33  & 5272          & \bf{0.00} & 4.46  & 5272  & \bf{0.00}  & 0.94  & 5214     & 1.10 & 1.48  & 5218     & 1.02 & 3.97  \\
rat195   & 6195  & 6174           & \bf{0.34}  & 2.22  & 6086          & 1.76 & 3.06  & 6139  & 0.90  & 2.00  & 6127     & 1.10 & 3.00  & 6125     & 1.13 & 6.46  \\
d198     & 6320  & 5985           & 5.30  & 1.86  & 6162          & 2.50 & 5.86  & 6290  & \bf{0.47}  & 7.14  & 6258     & 0.98 & 2.71  & 6212     & 1.71 & 5.46  \\
kroA200  & 6123  & 6048           & 1.22  & 2.73  & 6084          & 0.64 & 4.64  & 6114  & \bf{0.15}  & 1.72  & 5971     & 2.48 & 2.39  & 6028     & 1.55 & 7.14  \\
kroB200  & 6266  & 6251           & \bf{0.24}  & 2.79  & 6190          & 1.21 & 5.46  & 6213  & 0.85  & 1.77  & 6163     & 1.64 & 3.17  & 6226     & 0.64 & 4.78  \\
gr202    & 8616  & 8111           & 5.86  & 2.05  & 8419          & 2.29 & 9.12  & 8605  & \bf{0.13}  & 10.45 & 8464     & 1.76 & 3.79  & 8542     & 0.86 & 9.68  \\
ts225    & 7575  & 7149           & 5.62  & 1.47  & 7510          & 0.86 & 6.15  & 7575  & \bf{0.00}  & 1.14  & 7486     & 1.17 & 3.12  & 7575     & \bf{0.00} & 9.80  \\
tsp225   & 7740  & 7353           & 5.00  & 2.38  & 7565          & 2.26 & 5.04  & 7488  & 3.26  & 2.58  & 7607     & 1.72 & 3.74  & 7681     & \bf{0.76} & 11.50 \\
pr226    & 6993  & 6652           & 4.88  & 1.97  & 6964          & \bf{0.41} & 15.50 & 6908  & 1.22  & 8.01  & 6950     & 0.61 & 3.37  & 6937     & 0.80 & 6.10  \\
gr229    & 6328  & 6190           & 2.18  & 4.42  & 6205          & 1.94 & 9.03  & 6297  & \bf{0.49}  & 11.65 & 6135     & 3.05 & 5.98  & 6197     & 2.07 & 17.40 \\
gil262   & 9246  & 8915           & 3.58  & 5.68  & 8922          & 3.50 & 6.07  & 9094  & 1.64  & 3.94  & 9090     & 1.69 & 7.57  & 9116     & \bf{1.41} & 22.60 \\
pr264    & 8137  & 7820           & 3.90  & 3.98  & 7959          & 2.19 & 17.88 & 8068  & 0.85  & 3.62  & 8118     & \bf{0.23} & 4.21  & 8086     & 0.63 & 7.44  \\
a280     & 9774  & 8719           & 10.79 & 4.53  & 9426          & 3.56 & 9.42  & 8684  & 11.15 & 3.22  & 9609     & \bf{1.69} & 4.83  & 9587     & 1.91 & 13.55 \\
pr299    & 10343 & 10305          & \bf{0.37}  & 6.07  & 10033         & 3.00 & 19.61 & 9959  & 3.71  & 3.95  & 10227    & 1.12 & 5.40  & 10254    & 0.86 & 16.01 \\
lin318   & 10368 & 9909           & 4.43  & 7.57  & 9758          & 5.88 & 12.18 & 10273 & \bf{0.92}  & 6.33  & 10155    & 2.05 & 9.05  & 10271    & 0.94 & 21.95 \\
rd400    & 13223 & 12828          & 2.99  & 14.49 & 12678         & 4.12 & 16.46 & 13088 & \bf{1.02}  & 7.74  & 12552    & 5.07 & 12.15 & 12848    & 2.84 & 53.39 \\\midrule
average  &  4724 &  4601  & 1.69  & 1.80  & 4646 & 0.96 & 4.18  &  4661  & 0.90  & 2.31  &  4661  & 0.88 & 2.19  & 4683  & \bf{0.58} & 5.90  \\\bottomrule
\end{tabular}
}
\end{table}

\section{Conclusion}\label{Section: Conclusion}
We have proposed a new meta-heuristic called ML-ACO that integrates machine learning (ML) with ant colony optimization (ACO) to solve the orienteering problem. Our ML model trained on optimally-solved problem instances, is able to predict which edges in the graph of a test problem instance are more likely to be part of the optimal route. We incorporated the ML predictions into the probabilistic model of ACO to bias its sampling towards using the predicted `high-quality' edges more often when constructing solutions. This in turn significantly boosted the performance of ACO in finding high-quality solutions for a test problem instance. We tested three different classification algorithms, and the experimental results showed that all of the ML-enhanced ACO variants significantly improved the classic ACO in terms of both speed of convergence and quality of the final solution. Of the three ML models, the SVM based predictions produced the best results for this application. The best results were obtained by using the prediction to modify the heuristic weights rather than just for the initial pheromone matrix. Importantly, our ML model trained on small synthetic problem instances generalized very well to large synthetic and real-world problem instances.


We see great potential of the integration between ML (more specifically  solution prediction) and meta-heuristics, and a lot of opportunities for future work.  First, there is a large family of meta-heuristics, that can potentially be improved by solution prediction.  Second, it would be interesting to see if this integrated technique also works on other combinatorial optimization problems as well as continuous, dynamic or multi-objective optimization problems. In particular, we expect this integrated technique would be more effective in solving a dynamic problem where the optimal solution changes over time. Based on the results shown in this paper, solution prediction is very greedy, and therefore can potentially adapt quickly to any changes occurring  in a problem. Third, there is a  large number of ML algorithms that can be used for solution prediction. Meta-heuristics will certainly benefit more from this type of hybridization, if we can further improve the accuracy of solution prediction. This paper shows that SVM, one of the simpler ML models, is already highly effective. However, given the large number of advanced ML methods developed in recent years, there may be others that are even more effective in this context of boosting meta-heuristics.

\setcounter{section}{0}
\renewcommand\thesection{\Alph{section}}

\section{Supplementary Methodology}

\subsection{A random sampling method for the orienteering problem}\label{Subsection: Random Sampling Method for Orienteering Problem}

Consider an orienteering problem instance $G(V,E,S,C)$ with a given time budget $T_\mathrm{max}$. Without loss of generality, we assume $v_1$ is the starting vertex and $v_n$ is the ending vertex. The main steps of our random sampling method to generate one feasible solution (route) are:
\begin{enumerate}
	\item Initialize a route with the starting vertex $v_1$;  
	\item Generate a random permutation of the candidate vertices $\{v_2,\ldots,v_{n-1}\}$ that can be visited;
	\item Consider the vertices in the generated permutation one by one, and add the vertices to the sample route which does not violate the time budget constraint;
	\item Add the ending vertex $v_n$ to the sample route.
\end{enumerate}
The pseudocode of the random sampling method is presented in Algorithm \ref{Algorithm: Random Sampling Methods}. It is obvious that the time complexity of generating one sample route by using this method is $\mathcal{O}(n)$, where $n$ is the number of vertices in a problem instance. Hence, the total time complexity of generating $m$ sample routes is $\mathcal{O}(mn)$. Furthermore, the sample size $m$ should be larger than $n$; otherwise there will be some edges that are never sampled. This is because the  number of edges in the directed complete graph is $n(n-1)$, and the total number of edges in $m$ sample routes is no more than $mn$. Therefore, each edge is expected to be sampled no more than $m/(n-1)$ times.

\begin{algorithm}[tb]
	\caption{\textsc{Random Sampling Method}}
	\label{Algorithm: Random Sampling Methods}
	\begin{algorithmic}[1]
		\Require vertex set $V$, vertex score set $S$, edge cost set $C$, time budget $T_\mathrm{max}$, number of samples to generate $m$.
		\For{$k$ from $1$ to $m$}
		\State Initialize the route $P^k$ with the starting vertex $v_1$;
		\State Initialize the object value $y^k \leftarrow S[v_1]$;
		\State Initialize the current vertex $v_{c} \leftarrow v_1$;
		\State Initialize the time used so far $t_{c} \leftarrow 0$;
		\State Generate a random permutation of $\{v_2,\ldots,v_{n-1}\}$;
		\For{$v_j$ in the generated random permutation}
		\If{$t_{c} + C[v_c,v_j] + C[v_j, v_n] \le T_\mathrm{max}$}
		\State Add $v_j$ to the route $P_k$;
		\State Update $y^k \leftarrow y^k + S[v_j]$; 
		\State Update $t_c \leftarrow t_c + C[v_c,v_j]$; 
		\State Update $v_c \leftarrow v_j$;		
		\EndIf
		\EndFor
		\State Add $v_n$ to the route $P_k$;
		\State Update $y^k \leftarrow y^k + S[v_n]$; 
		\EndFor \\ 					
		\Return $\{P^1,\ldots,P^{m}\}$ and $\{y^1,\ldots,y^{m}\}$.
	\end{algorithmic}
\end{algorithm}

\subsection{An efficient method for computing the statistical measures}\label{Subsection: Computing Statistical Measures Efficiently}

In the main paper, we used a binary string $\bm{x}$ to represent a sample solution (route), where $x_{i,j} = 1$ if the edge $e_{i,j}$ is in the route; otherwise $x_{i,j} = 0$. We have shown that directly computing the ranking-based measure and correlation-based measure based on the binary string representation $\bf{x}$ costs $\mathcal{O}(mn^2)$. Here, we adapt the method proposed in \citep{sun2019using} to efficiently compute the two statistical measures based on set representation $P$, which only stores the edges appearing in the corresponding sample route. 

Let $\{P^1,\ldots,P^{m}\}$ be the set representation of  the $m$ randomly generated solutions; $\{\bm{x}^1,\ldots,\bm{x}^m\}$ be the corresponding binary string representation; and $\{y^1,\ldots,y^m\}$ be their objective values. Because $x_{i,j}^k$ are binary variables, we can simplify the calculation of Pearson correlation coefficient by using the following two equalities:
\begin{equation}
\sum_{k=1}^{m}(x_{i,j}^k-\bar{x}_{i,j})^2 = \bar{x}_{i,j}(1-\bar{x}_{i,j})m,
\end{equation}
\begin{equation}
\sum_{k=1}^{m}(x_{i,j}^k - \bar{x}_{i,j})(y^k - \bar{y})  = (1- \bar{x}_{i,j}) s_{i,j}^1 - \bar{x}_{i,j} s_{i,j}^0,
\end{equation}
where $\bar{x}_{i,j} = \sum_{k= 1}^{m}x_{i,j}^k/m$,  $\bar{y} = \sum_{k=1}^{m}y^k/m$ and 
\begin{equation}
s_{i,j}^1 = \mathop{\sum\limits_{1\le k \le m}}\limits_{x_{i,j}^k = 1}(y^k - \bar{y}); \; \text{and} \; s_{i,j}^0 = \mathop{\sum\limits_{1\le k \le m}}\limits_{x_{i,j}^k = 0}(y^k - \bar{y}). 
\end{equation}
The proof of these two equalities can be found in \citep{sun2019using}. We then are able to compute the two statistical measures in $\mathcal{O}(mn+n^2)$ by using Algorithm \ref{Algorithm: Statistical Measures}. Computing our ranking-based measure $f_r$ based on the set representation is straightforward, i.e., scanning through the edges in each sample route $P$ to accumulate the rankings. To compute the correlation-based measure, we first iterate through the edges in each sample route $P$ to accumulate  $\bar{x}_{i,j}$ and $s_{i,j}^1$, i.e., line \ref{line: for1} to \ref{line: for2} in Algorithm \ref{Algorithm: Statistical Measures}. Our correlation-based measure $f_c$ can then be easily computed based on $\bar{x}_{i,j}$ and $s_{i,j}^1$ (line \ref{line: for3} to \ref{line: for4} in Algorithm \ref{Algorithm: Statistical Measures}).

\begin{algorithm}[tb]
	\caption{\textsc{Computing Statistical Measures}}
	\label{Algorithm: Statistical Measures}
	\begin{algorithmic}[1]
		\Require samples $\mathbb{P}$, objective values $Y$, number of samples $m$, number of vertices $n$, and edge set $E$.
		\State Sort the samples in $\mathbb{P}$ based on objective value $Y$; and use $r^k$ to denote the ranking of $k^\mathrm{th}$ sample $P^k$;
		\State Compute mean objective value: $\bar{y} \leftarrow \sum_{k=1}^{m}y^k/m$;
		\State Compute objective difference: $y_d \leftarrow \sum_{k=1}^{m}(y^k - \bar{y})$; 
		\State Compute objective variance: $\sigma_y \leftarrow \sum_{k=1}^{m}(y^k - \bar{y})^2$; 
		\State Initialize $f_r$, $\bar{x}_{i,j}$ and $s_{i,j}^1$ to $0$,  for each $e_{i,j} \in E$; 					
		\For{$k$ from $1$ to $m$}\label{line: for1}
		\For{$idx$ from $1$ to $|P^k|-1$}		
		\State $i \leftarrow P^k[idx]$, $j \leftarrow P^k[idx+1]$;
		\State $f_r(e_{i,j}) \leftarrow f_r(e_{i,j}) + 1/r^k$;
		\State $\bar{x}_{i,j} \leftarrow \bar{x}_{i,j} + 1 / m$;
		\State $s_{i,j}^1 \leftarrow s_{i,j}^1 + (y^k - \bar{y})$;
		\EndFor
		\EndFor\label{line: for2}			
		\For{$i$ from $1$ to $n$}\label{line: for3}	
		\For{$j$ from $1$ to $n$ and $j\ne i$}
		\State $\sigma_{c_{i,j}} \leftarrow (1-\bar{x}_{i,j})s_{i,j}^1 - \bar{x}_{i,j}(y_d - s_{i,j}^1)$;
		\State $\sigma_{x_{i,j}}  \leftarrow \bar{x}_{i,j}(1-\bar{x}_{i,j})m$;
		\State $f_c(e_{i,j}) \leftarrow  \sigma_{c_{i,j}}/ \sqrt{\sigma_{x_{i,j}}\sigma_y}$; 
		\EndFor
		\EndFor\label{line: for4}	\\
		\Return $f_r$ and $f_c$. 
	\end{algorithmic}
\end{algorithm}

\subsection{A brief introduction of the machine learning algorithms used }\label{Subsection: ml algorithms} 

\textbf{Support Vector Machine} (SVM): Given a training set $\mathbb{S}=\{(\bm{f}^i,l^i)\,|\, i = 1,\ldots,n_t\}$, the aim of SVM is to find a decision boundary ($\bm{w}^T\bm{f}+b = 0$)  in the feature space to maximize the so-called geometric margin, defined as the smallest distance from a training point to the decision boundary  \citep{boser1992training,cortes1995support}. We use an L2-regularized linear SVM model, that finds the optimal decision boundary by solving the following quadratic programming with linear constraints:
\begin{align}
	\min_{\bm{w},b, \bm{\xi}} & \quad \frac{1}{2}\bm{w}^T\bm{w} + r^+ \sum_{l^i=1}\xi^i + r^- \sum_{l^i=-1}\xi^i, \label{eq:svm} \\ 
	s.t. & \quad l^i\big(\bm{w}^T\bm{f}^i+b\big) \ge 1 - \xi^i, \quad i = 1, \ldots n_t,\\
	& \quad \xi^i \ge 0, \quad i = 1, \ldots n_t,
\end{align}
where  $r^+>0$ and $r^->0$ are the regularization parameters for positive and negative training points; and $\xi^i$, $i=1,\ldots,n_t$ are slack variables.   

\textbf{Logistic Regression} (LR) uses a loss function derived from the logistic function $g(x) = 1/(1+e^{-x})$, whose output is in between $[0,1]$ and can be interpreted as probability. LR aims to separate positive and negative training points by maximum likelihood estimation \citep{bishop2006pattern}. We use an L2-regularized LR model that fits its parameters ($\bm{w}$, $b$) by solving the following optimization problem:
\begin{equation}
	\min_{\bm{w},b} \quad  \frac{1}{2}\bm{w}^T\bm{w}  + r^+ \sum_{l^i=1}  \log_2(1+e^{-\bm{w}^T\bm{f}^i-b}) 
	 + r^- \sum_{l^i=-1}  \log_2(1+e^{\bm{w}^T\bm{f}^i+b}). \label{eq:lr}
\end{equation}

\textbf{Graph Convolutional Network} (GCN) is a convolutional neural network that makes use of graph structure when classifying vertices in a graph \citep{kipf2016semi}. Consider a simple GCN model with only two layers: the input layer contains  feature vectors $\bm{f}$ and the output layer is a predicted scalar  $z$ for a vertex in a graph.  To compute $z_i$ for vertex $v_i$ in a graph, GCN aggregates its feature vector $\bm{f}^i$ with that of its neighbours $\mathcal{N}_i$:
\begin{equation}
z_i = \bm{w}_0 \bm{f}^i + \bm{w}_1 \sum_{v_j \in \mathcal{N}_i}  \sqrt{d_i  d_j} \bm{f}^j,
\end{equation}
where $d_i$ and $d_j$ are the degrees of vertex $v_i$ and $v_j$; $\bm{w}_0$ and $\bm{w}_1$ are the weights to be optimized.
This two-layer GCN model is a simple linear classifier, which is not expected to work well in practice. Thus, we usually use multiple hidden layers between the input and output layers, and each hidden layer can have multiple `neurons'. A hidden layer basically takes the output of its previous layer as input, and performs a linear transformation of its input. The intermediate output of the linear transformation is then filtered by an activation function to make GCN a non-linear classifier. In our experiments, the GCN model consists of 20 layers and each hidden layer has 32 neurons. The activation function used is the ReLU function \citep{nair2010rectified}, defined as $\text{ReLU}(x) = \max(0, x)$. The weights ($\bm{w}$) of GCN are optimized via stochastic gradient descent with L2-regularized cross-entropy loss function  \citep{kipf2016semi}. In the case of binary classification, the cross-entropy loss function is identical to the loss function of the LR algorithm:
\begin{equation}
\min_{\bm{w}} \quad  \frac{1}{2}\bm{w}^T\bm{w}  + r^+ \sum_{l^i=1}  \log_2(1+e^{-z_i})  + r^- \sum_{l^i=-1}  \log_2(1+e^{z_i}), \label{eq:gcn}
\end{equation}
where $\bm{w}$ is a vector of all GCN's weights to be optimized, and $z_i$ is the output (prediction) of GCN for the $i^\mathrm{th}$ training point. Because our training points are edges instead of vertices in the graphs of the orienteering problem instances, the GCN model cannot be directly applied to make predictions for edges in the graphs. To tackle this, we transfer the original graph ($G$) to its line graph ($\bar{G}$) such that the edges in $G$ are now vertices in $\bar{G}$ and the neighbouring edges (i.e., edges sharing a common vertex)  in $G$  are now neighbouring vertices (i.e., vertices sharing a common edge)  in $\bar{G}$. We then can apply the GCN model on the line graph $\bar{G}$ to make predictions for the edges in the original graph $G$.

\section{Adapting ML-ACO to Solve the Maximum Weighted Clique Problem}
\label{sec::mwcp}

As our ML-ACO algorithm is a generic approach, we apply it to solve another combinatorial optimization problem, the maximum weighted clique problem (MWCP). The MWCP is a variant of the maximum clique problem, which is a fundamental problem in graph theory with a wide range of real-world applications \citep{wu2015review,malladi2017clustered,letchford2020stable,blum2021solving}. Solving the MWCP is NP-hard, and a large number of solution methods have been developed for this problem recently. These include exact solvers such as branch-and-bound algorithms \citep{jiang2017exact,jiang2018two,li2018new,san2019new} and local search methods \citep{wang2016two,cai2016fast,zhou2017push,nogueira2018cpu,wang2020sccwalk}. 

Here, the aim of our ML model is to predict the `probability' of a vertex being part of the maximum weighted clique. To train our ML model, we construct a training set using eighteen optimally-solved small graphs ($|V| < 1000$) from the standard DIMACS library. The original DIMACS graphs are unweighted, and thus we assign a weight $w_i = (i \mod 200 )+ 1$ to the vertex $v_i$  $(i = 1,\ldots,|V|)$, following the previous works \citep{wang2016two,cai2016fast,jiang2017exact,jiang2018two}. Each training instance corresponds to a vertex in a training graph. We extract six features to characterize a vertex, including graph density, vertex weight, vertex degree, an upper bound and two statistical features described in  \citep{sun2019using}. A training instance is labeled as $1$ if the corresponding vertex belongs to the maximum weighted clique; otherwise it is labeled as $-1$. We then train a linear SVM to classify whether a vertex belongs to the maximum weighted clique or not. 

We use fifteen larger graphs ($|V| \ge 1000$) from the DIMACS library as our test problem instances. For each problem instance, we use the trained ML model to predict a probability value $p_i$ for each vertex $v_i  \in V$. The predicted values ($p_i$) are then incorporated into the MMAS algorithm to guide its sampling towards larger-weighted cliques. More specifically, we use $p_i$ to set the heuristic weight: $\eta_{i} = p_i \cdot w_i$, for each $v_i \in V$. The parameter settings for MMAS and linear SVM are the same as before. We compare our ML-ACO algorithm against two exact solvers -- TSM \citep{jiang2018two} and WLMC \citep{jiang2017exact} as well as two heuristic methods LSCC \citep{wang2016two} and FastWClq \citep{cai2016fast} for solving the MWCP. The cutoff time for each algorithm is set to 1000 seconds.  

The best objective values obtained by each algorithm in 25 independent runs are presented in Table~\ref{tab:mwcp}. We can observe that our ML-ACO algorithm is comparable to the state-of-the-art algorithms for solving the MWCP. On average, the best objective values found by our ML-ACO algorithm are significantly better than those found by FastWClq, WLMC and TSM. The LSCC algorithm performs the best and generates slightly better results than our ML-ACO algorithm. These results are interesting, because generic solution methods such as our ML-ACO algorithm are not often expected to be as competitive as specialized solvers.

\begin{table}[!t]
	\centering
 	\small
	\caption{The best objective values generated by our ML-ACO algorithm and four state-of-the-arts for solving the MWCP. The best results are highlighted in bold.}
	\label{tab:mwcp}
	\begin{tabular}{lrrrrrr}
		\toprule
		Graph & $|V|$ & ML-ACO & FastWClq & LSCC &  WLMC & TSM \\\midrule		
		p\_hat1000-1 & 1000 & 1514  & 1514      & 1514   & 1514     & 1514    \\
		p\_hat1000-2 & 1000  & 5777  & 5777      & 5777   & 5777     & 5777    \\
		p\_hat1000-3 & 1000  & \bf{8111}  & 8058      & \bf{8111}   & 8076     & \bf{8111}    \\
		p\_hat1500-1 & 1500  & 1619  & 1619      & 1619   & 1619     & 1619    \\
		p\_hat1500-2 & 1500  & \bf{7360}  & 7327      & \bf{7360}   & \bf{7360}     & \bf{7360}    \\
		p\_hat1500-3 & 1500  & \bf{10321} & 10057     & \bf{10321}  & 9846     & 10119   \\
		DSJC1000.5 & 1000  & 2186  & 2186      & 2186   & 2186     & 2186    \\
		san1000 & 1000  & 1716  & 1716      & 1716   & 1716     & 1716    \\
		C1000.9 & 1000  & 9191  & 8685      & \bf{9254}   & 7317     & 7341    \\
		C2000.5 & 2000  & \bf{2466}  & \bf{2466}      & \bf{2466}   & 2360     & 2407    \\
		C2000.9 & 2000  & 10888 & 9943      & \bf{10964}  & 7738     & 8228    \\
		C4000.5 & 4000  & 2776  & 2645      & \bf{2792}   & 2383     & 2402    \\
		MANN\_a45 & 1035  & 34209 & 34111     & 34243  & \bf{34265}    & 34259   \\
		hamming10-2 & 1024  & 50512 & 50512     & 50512  & 50512    & 50512   \\
		hamming10-4 & 1024  & 5127  & 4982      & \bf{5129}   & 4738     & 4812    \\\midrule
		average & - & 10252 & 10107     & 10264  & 9827   & 9891  \\\bottomrule
	\end{tabular}
\end{table}

\section*{Acknowledgment}
This work was supported by an ARC (Australian Research Council) Discovery Grant (DP180101170).

\bibliography{MLACO.bib}

\end{document}